\documentclass[sigconf]{acmart}
\usepackage[utf8]{inputenc} 
\usepackage[T1]{fontenc}    
\usepackage{hyperref}       
\usepackage{url}            
\usepackage{booktabs}       
\usepackage{amsfonts}       
\usepackage{bbm}
\usepackage{nicefrac}       
\usepackage{microtype}      
\usepackage{xcolor}         
\usepackage{amsmath}
\usepackage{appendix}
\usepackage{multirow}
\usepackage{array,float}
\usepackage{threeparttable}
\usepackage{wrapfig}
\usepackage{graphicx}

\newcommand{\ModelName}{{GASSER}}

\newcommand{\pn}{$\pm$}

\usepackage{paralist}
\usepackage{mathtools}

\usepackage{amsthm}
\usepackage{color}

\newcommand{\cut}[1]{{}}

\newcommand{\cB}{{\mathcal{B}}}

\newcommand{\cE}{{\mathcal{E}}}

\newcommand{\cL}{{\mathcal{L}}}

\newcommand{\cO}{{\mathcal{O}}}

\newcommand{\bB}{\boldsymbol{\cB}}

\makeatletter
\let\@@span\span
\def\sp@n{\@@span\omit\advance\@multicnt\m@ne}
\makeatother

\newcommand{\bc}{\begin{center}}
\newcommand{\ec}{\end{center}}

\newcommand{\bdm}{\begin{displaymath}}
\newcommand{\edm}{\end{displaymath}}

\newcommand{\beq}{\begin{equation}}
\newcommand{\eeq}{\end{equation}}

\newcommand{\bfl}{\begin{flushleft}}
\newcommand{\efl}{\end{flushleft}}

\newcommand{\bt}{\begin{tabbing}\vspace{-1em}}
\newcommand{\et}{\vspace{-1em}\end{tabbing}}

\newcommand{\beqn}{\begin{align}\vspace{-1em}}
\newcommand{\eeqn}{\vspace{-1em}\end{align}}
\newcommand{\bueqn}{\begin{align*}\vspace{-1em}} 
\newcommand{\eueqn}{\vspace{-1em}\end{align*}} 

\newcommand{\beqs}{\begin{equation*}\vspace{-1em}} 
\newcommand{\eeqs}{\vspace{-1em}\end{equation*}}

\newtheorem{theorem}{Theorem}

\AtBeginDocument{%
 }

\copyrightyear{2024}
\acmYear{2024}
\setcopyright{rightsretained}
\acmConference[CIKM '24]{Proceedings of the 33rd ACM International Conference on Information and Knowledge Management}{October 21--25, 2024}{Boise, ID, USA}
\acmBooktitle{Proceedings of the 33rd ACM International Conference on Information and Knowledge Management (CIKM '24), October 21--25, 2024, Boise, ID, USA}
\acmDOI{10.1145/3627673.3679762}
\acmISBN{979-8-4007-0436-9/24/10}

\makeatletter
\gdef\@copyrightpermission{
  \begin{minipage}{0.3\columnwidth}
   \href{https://creativecommons.org/licenses/by/4.0/}{\includegraphics[width=0.90\textwidth]{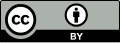}}
  \end{minipage}\hfill
  \begin{minipage}{0.7\columnwidth}
   \href{https://creativecommons.org/licenses/by/4.0/}{This work is licensed under a Creative Commons
Attribution International 4.0 License.}
  \end{minipage}
  \vspace{5pt}
}
\makeatother

\begin{document}

\title{Spectral-Aware Augmentation for Enhanced Graph Representation Learning}
\author{Kaiqi Yang}
\orcid{0009-0008-1719-1479}
\affiliation{%
  \institution{Michigan State University}
  \city{East Lansing}
  \state{MI}
  \country{USA}
}
\email{kqyang@msu.edu}

\author{Haoyu Han}
\orcid{0000-0002-2529-6042}
\affiliation{%
  \institution{Michigan State University}
  \city{East Lansing}
  \state{MI}
  \country{USA}
}
\email{hanhaoy1@msu.edu}

\author{Wei Jin}
\orcid{0000-0002-5054-954X}
\affiliation{%
  \institution{Emory University}
  \city{Atlanta}
  \state{GA}
  \country{USA}
}
\email{wei.jin@emory.edu}

\author{Hui Liu}
\authornote{Corresponding author.}
\orcid{0000-0002-3555-3495}
\affiliation{%
  \institution{Michigan State University}
  \city{East Lansing}
  \state{MI}
  \country{USA}
}
\email{liuhui7@msu.edu}

\begin{abstract}
Graph Contrastive Learning (GCL) has demonstrated remarkable effectiveness in learning representations on graphs in recent years. To generate ideal augmentation views, the augmentation generation methods should preserve essential information while discarding less relevant details for downstream tasks. 
However, current augmentation methods usually involve random topology corruption in the spatial domain, which fails to adequately address information spread across different frequencies in the spectral domain. Our preliminary study highlights this issue, demonstrating that spatial random perturbations impact all frequency bands almost uniformly. Given that task-relevant information typically resides in specific spectral regions that vary across graphs, this one-size-fits-all approach can pose challenges. We argue that indiscriminate spatial random perturbation might unintentionally weaken task-relevant information, reducing its effectiveness.

To tackle this challenge, we propose applying perturbations selectively, focusing on information specific to different frequencies across diverse graphs. In this paper, we present \textbf{\ModelName}, a model that applies tailored perturbations to specific frequencies of graph structures in the spectral domain, guided by spectral hints. Through extensive experimentation and theoretical analysis, we demonstrate that the augmentation views generated by \textbf{\ModelName} are adaptive, controllable, and intuitively aligned with the homophily ratios and spectrum of graph structures.
\end{abstract}

\ccsdesc[500]{Computing methodologies~Machine learning algorithms}
\ccsdesc[500]{Computing methodologies~Artificial intelligence}

\keywords{Graph Contrastive Learning, Graph Neural Networks, Graph Spectrum}


\maketitle

\section{Introduction}

Graphs serve as effective tools for characterizing relationships between entities in numerous domains such as social media~\cite{fan2019deep,battaglia2018relational}, chemistry~\cite{ye2020symmetrical}, biology~\cite{li2021graph}, and finance~\cite{wang2021review,fan2019graph}. To facilitate the graph-related tasks, graph neural networks (GNNs)~\cite{zhou2020graph,jin2020graph,xu2018powerful} have been developed to extract useful information from graph data. Despite their promise, GNNs have mainly been studied in the context of supervised end-to-end training, which demands sufficient labeled data. To address this challenge, self-supervised learning approaches have been proposed ~\cite{RN207,RN134,dgi,RN202,RN202}, which extend from the image domain to the graph domain by designing pretext tasks to derive self-supervision for unlabeled data.

Among various self-supervised learning techniques, graph contrastive learning (GCL)~\cite{RN55,robinson2020contrastive,mvgrl,bielak2022graph,gcc} has attracted significant attention due to its remarkable benefits to downstream tasks. 
Most GCL methods first generate several views of the graph data by performing data augmentation and then learn graph representations that are invariant to the augmentation views~\cite{RN55,liu2022graph} by maximizing mutual information between these views. 
Recent studies~\cite{tsai2020self,tian2020makes,arora2019theoretical} have attributed the success of contrastive learning to the preservation of task-relevant information in the augmentation views, which highlights the importance of crafting good views that maintain a sufficient amount of task-relevant information. 

To obtain the augmentation views for graph data, the majority of existing GCL methods~\cite{RN55,robinson2020contrastive,mvgrl,bielak2022graph,gcc} perform spatial data augmentation on the graph topology such as random node deletion~\cite{RN55} and edge insertion~\cite{liu2022graph}. However, these methods perform the augmentations on the spatial domain while ignoring their influence on properties in the spectral domain. As demonstrated in recent literature~\cite{entezari2020all,jin2020graph}, modification to the structure may essentially perturb the spectral properties of the graph and lead to degraded performance. 
Moreover, these approaches overlook the fact that different types of graphs have different requirements for good augmentation views. 

For example, the distribution of task-relevant information at different frequency bands differs between graphs with distinct homophily~\cite{zhu2020beyond}. 
In Section~\ref{sec:pre}, we further verify  that the task-relevant information of graphs tends to be concentrated in a particular region of the graph spectrum and such regions vary between homophilic and heterophilic graphs.
Consequently, applying the same augmentation without adapting to different types of graphs could lead to sub-optimal augmentation views. 
In addition, we observe that the random spatial augmentations on graph structure have a nearly uniform effect on the low-, middle-, and high-frequency bands. This uniform change across the spectrum may perturb the informative frequency bands, leading to reduced task-relevant information. These motivate us to design new augmentation strategies that consider the graph's spectral characteristics.

To generate effective augmentation views {of graph data for node classification task}, we argue that a spectral-aware augmentation strategy is desired for GCL. In this work, we propose \underline{G}r\underline{A}ph contrastive learning with \underline{S}elective \underline{S}pectrum p\underline{ER}turbation, namely \ModelName{}, to augment the graph data by altering the graph spectrum selectively. Specifically, we inject noises into a selected group of eigenvectors acquired from the graph adjacency matrix. By selecting an appropriate subset of eigenvectors for augmentation, \ModelName{} can provide the flexibility to adapt the augmentation process to benefit different types of graphs, especially those with different homophily ratios.

Our main contributions are summarized as follows:

\vspace{-3pt}
\begin{itemize}
 \item We empirically show that the spatial augmentation method could cause spectral changes in a nearly uniform manner across different frequency bands. 
\item We propose a GCL framework \ModelName{} with a new augmentation method, which carefully modifies the graph data from the spectral domain aiming to preserve the graph's key information.

\item We provide theoretical analysis to motivate our proposed method and justify its efficacy. 

\item We conduct extensive experiments {on eight node-classification benchmarks of different homophily ratios}, demonstrating the effectiveness of \ModelName{} on various real-world datasets. Moreover, we showcase that the proposed method can be seamlessly integrated into various existing contrastive learning frameworks and it exhibits superior robustness. 
\end{itemize}

\vspace{-13pt}
\section{Related Work}

\label{work:principlecl}
Contrastive learning is a family of self-supervised learning algorithms in which the models are learned by comparing the input samples~\cite{grlsurvey,hamilton2017representation,perozzi2014deepwalk}. To support the contrastive mechanism, positive (similar) and negative (dissimilar) samples are required as substitutes for supervision in most cases~\cite{RN254}. As stated in~\cite{gca}, the intrinsic semantics and attributes should be preserved across the original view and augmentation views, thus making model learning invariant information by maximizing the agreement of positive pairs. Beyond that, some works intend to understand contrastive learning theoretically or empirically. For example, by projecting representations on the unit hypersphere, the contrastive representation learning with \textit{alignment} and \textit{uniformity} is measured in~\cite{wang2020understanding}, indicating that the desirable representation should push the learned features of positive pairs close, as well as the distribution of learned features uniform and thus informative. 

The design of augmentations plays a vital role in contrastive learning. Researchers on graph contrastive learning are devoted to designing better augmentations for graphs~\cite{suresh2021adversarial,zhang2022costa,auggraph,gca,sun2023graph}. As one of the topology augmentations for graph data~\cite{trivedi2022augmentations}, edge perturbation including edge adding and dropping is always deployed in a spatial and random way. Nevertheless, its assumption of the uniform distribution of edges has been shown to be plausible or even incorrect~\cite{chang2021notall,afgcl}. Recently some works design selective edge perturbation by spatial properties and methods, such as node centrality~\cite{gca}, node similarity~\cite{chi2022enhancing}, and graph edit distance~\cite{kim2022graph}. Some other works perturb the graph structure globally, such as graph diffusion~\cite{mvgrl} and random-walk~\cite{gcc,tong2006fast}. However, the spatial perturbation processes the overall graph with high overhead and still suffers from changing semantics potentially~\cite{lee2022augmentation}. 

There are several works on the graph spectrum to construct effective augmentation views. From the perspective of the graph spectrum, different parts of the graph spectrum indicate different structure properties~\cite{chung1997spectral}, while it is ignored by random spatial augmentation~\cite{wang2022can}. GCL-SPAN~\cite{span} constructs augmentation graphs with the maximized difference of graph spectrum by flipping the edges selectively, and SGCL~\cite{amur2023sagcl} proposes the spectral graph cropping augmentation based on the eigenvector corresponding to the second smallest eigenvalue. There are also works concentrating on the spectrum of feature maps. For example, SFA~\cite{sfa} tries to rebalance the spectrum and align the spectral features of augmentation views. However, searching for spectral augmentation usually takes a feasible set consisting of all the possible graphs (feature maps) and their spectrum, which is extremely complicated and time-consuming. 

\textbf{Comparisoin with Existing Spectrum-based GCL.} Our proposed method enjoys the following advantages -- (i) versatility: our method generally achieves better performance on both homophily and heterophily graphs; (ii) efficiency: our method removes the need for learning structure for augmentation view, resulting in significantly improved efficiency compared to GCL-SPAN~\cite{span}; and (iii) robustness: our method learns invariant node representations to the noise in spectrum and can effectively handle spectrum changes caused by structure attacks.

\vspace{-14pt}
\section{Preliminary Study}

In this section, we perform preliminary studies to motivate the need for carefully crafting spectral augmentation. Before that, we first introduce key notations and definitions.

We denote a graph as $\mathcal G=(\mathcal V, \mathcal E, {\bf X})$, where $\mathcal V$ is the node set and $\mathcal E$ is the edge set, with the number of nodes $n=|\mathcal V|$ and the number of edges $m=|\mathcal E|$. ${\bf X}\in \mathbb R^{n\times d}$ is the feature matrix where $d$ is the feature dimension. The feature vector of node $i$ is the $i$-th row of ${\bf X}$ denoted as ${\bf x}_i$. The degree of a node $v_i$ is denoted as $d_{v_i}$, and the diagonal degree matrix is denoted as ${\bf D}$ with ${\bf D}_{i,i}=d_{v_i}$. The adjacency matrix of $\mathcal G$ is ${\bf A}\in\{0,1\}^{n\times n}$, in which the $ij$-{th} entry is 1 if and only if there is an edge between $v_i$ and $v_j$. We denote ${\bf A}_\text{sym}={\bf D}^{-1/2}{\bf A}{\bf D}^{-1/2}$ as the symmetric normalized version of the adjacency matrix. The Laplacian matrix and its normalized version are defined as ${\bf L}={\bf D}-{\bf A}$ and ${\bf L}_\text{sym}={\bf D}^{-1/2}{\bf L}{\bf D}^{-1/2}$, respectively. 
The eigen-decomposition of ${\bf A}_\text{sym}$ is denoted as ${\bf A}_\text{sym}=\bf{\Phi} {\bf\Omega} \bf{\Phi}^\top$, where $\bf{\Phi}$ is an orthogonal matrix with the eigenvectors ${\bf \phi}_i$ as the $i$-th column of $\bf{\Phi}$. The diagonal matrix ${\bf\Omega}$ consists of all the eigenvalues $\omega_i={\bf\Omega}_{i,i}$, ordered descendingly (i.e., $\omega_0\geq\omega_1\geq\dots\geq\omega_{n-1}$) to keep the corresponding relations with the eigenvalues of the Laplacian matrix~\cite{sagt}. The tuple $(\omega_i, {\bf \phi}_i)$ is an eigenpair of the decomposed matrix and ${\bf A}_\text{sym}[i]={\bf \phi}_i\omega_i{\bf \phi}_i^\top$ is defined as a \textit{decomposed component}. Similarly, for the Laplacian matrix, we have the following eigen-decomposition:  ${\bf L}_\text{sym}=\bf{U\Lambda U}^\top$ with $\lambda_0\leq\lambda_1\leq\dots\leq\lambda_{n-1}$, ($\lambda_i, u_i$) and ${\bf L}_\text{sym}[i]={\bf u}_i\lambda_i{\bf u}_i^\top$.

\vspace{-5pt}
\subsection{Task-Relevant Information from the Spectral Domain} \label{sec:pre}
For a given downstream task, the task-relevant information may be concentrated in different frequency bands for graphs of different properties. In this subsection, we use node classification as one example to illustrate that the distribution of task-relevant information can be distinct for graphs with different homophily~\cite{zhu2020beyond},  an important property for real-world graphs. 
\begin{align*}
h = \frac{|\{(v_i,v_j):(v_i,v_j)\in\mathcal E, {\bf y}_{v_i}={\bf y}_{v_j}\}|}{|\mathcal E|}
\end{align*}\vspace{-1em}

Homophily $h$ measures how similar the connected nodes are, and can be calculated as the fraction of intra-class edges in a graph~\cite{zhu2020beyond}.

Specifically, given a graph $\mathcal{G}$ and a label vector $\mathbf{y} \in \{0,1\}^{ n}$ defined on this graph, the homophily can be formulated as: 
$h(\mathcal{G}, {\bf y}) = \frac{1}{ m}\sum\nolimits_{(v_1,v_2)\in\mathcal{E}} \mathbbm{1}({\bf y}_{v_1} = {\bf y}_{v_2})$, 
where ${\bf y}_{v_i}$ indicates node $v_i$'s label and $\mathbbm{1}({\cdot})$ is the indicator function. Since the eigenvectors $\{{\bf u}_i\}$ of the unnormalized graph Laplacian ${\bf L}$ form a complete basis of $\mathbb{R}^{ n}$, the label vector $\mathbf{y}$ can be denoted as a linear combination of $\left\{\mathbf{u}_i\right\}$, i.e., $\mathbf{y}=\sum_{i=0}^{ n-1} c_{i} \mathbf{u}_i$ with  $c_{i}$ the coefficient of $\mathbf{y}$ at the $i$-th frequency component. The coefficients $\left\{c_{i}\right\}$ can be viewed as the spectrum of signal ${\bf y}$ w.r.t. the graph $\mathcal{G}$. Then, we present the following theorem:

\begin{theorem}
Given two label vectors ${\bf y}\in\{0,1\}^{ n}$ and ${\bf \hat{y}}\in\{0,1\}^{ n}$  which are defined on $\mathcal{G}$, we can decompose them into ${\bf y} = \sum_{i=0}^{ n-1} c_i{\bf u}_i$ and ${\bf \hat{y}} = \sum_{i=0}^{ n-1} \hat{c}_i{\bf u}_i$. We denote their homophily as $h_1$ and $h_2$, respectively. If the classes are balanced and $h_1-h_2=\Delta (\Delta>0)$, there exists an integer $M (0<M \leq n-1)$ such that $\sum_{i=M}^{ n-1}\hat{c}^2_i \geq  \sum_{i=M}^{ n-1}{c}^2_i + \frac{2\Delta  m}{\lambda_M  n} $ and $\sum_{i=0}^{M-1}\hat{c}^2_i\leq   \sum_{i=0}^{M-1}{c}^2_i - \frac{2\Delta  m}{\lambda_M  n}$. 
\label{thm:1}
\end{theorem}
The above theorem is an extension of the analysis in~\cite{zhu2020beyond}. It gives us a hint that for graphs with higher homophily, the label vector has a larger summation of spectral coefficients at low-frequency components and a smaller summation of coefficients at high-frequency components, compared to graphs with lower homophily. Moreover, the gap of the summation of coefficients $\frac{2\Delta  m}{\lambda_M  n}$ enlarges with the increase of homophily difference $\Delta$. It indicates that the distribution of task-relevant information at different frequency bands differs among graphs with different homophily.
Furthermore, to compare the spectral coefficients at different frequency bands within one graph, we introduce the following theorem. 
\begin{theorem}\label{thm:2} If the classes are balanced and $h_1 < 1- \frac{\lambda_\text{max} n}{8 m}$ with $\lambda_\text{max}$ the maximum eigenvalue of\ ${\bf L}$, there exists an integer $M' (0<M' \leq n-1)$ such that $\sum_{i=M'}^{ n-1} {c}^2_i > \sum_{i=0}^{M'-1}{c}^2_i$.
\end{theorem}
The detailed proof can be found in Section~\ref{app:proof2}. Theorem~\ref{thm:2} indicates that when the homophily is small, the coefficients at higher-frequency components are dominant. Similarly, when the homophily is large, the coefficients at lower-frequency components are dominant. It suggests that information is not evenly distributed across the graph spectrum.  
It is worth noting that in the extreme case where the homophily is $1$, the label vector ${\bf y}$ is only related to the first eigenvector ${\bf u}_0$ as proved at the end of Section~\ref{app:proof2}. This result further supports that there are task-relevant frequency bands and task-irrelevant ones, which motivates us to investigate whether the existing data augmentations can preserve the information in task-relevant frequency bands.

\begin{figure}
\begin{center}
\includegraphics[width=0.4\textwidth]{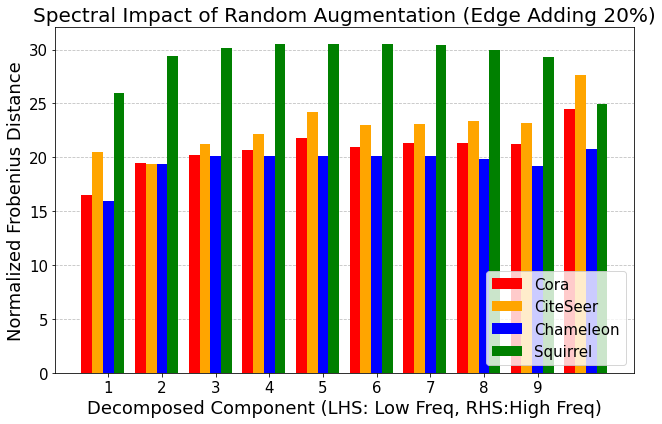}
\end{center}
\vspace{-1em}
\caption{Normalized distance between the grouped decomposed components of Laplacian matrixes of the original and augmented view with random 20\% edge insertion.} 
\label{fig:frob}
\vspace{-1em}
\end{figure}

\subsection{Impact of Random Spatial Augmentation on Graph Spectrum\label{spectraldifference}}
In the last subsection, we have demonstrated that the task-relevant information is not evenly distributed over the graph spectrum. In this subsection, we investigate how the commonly employed graph data augmentation, i.e., random spatial augmentation, affects different frequency bands. 
Following the settings in~\cite{afgcl}, we divide the Laplacian decomposed components  $\{{\bf L}_\text{sym}[i] | \ i=0,1,\ldots, n-1\}$ into 10 groups based their eigenvalues, and then add up all the decomposed components in each group: 

\vspace{-1em}\begin{align*}{\bf L}_\text{sym}^k=\sum\nolimits_{i\in[\frac{(k-1)N}{10},\frac{kN}{10})}{\bf L}_\text{sym}[i],\ k\in[1,10]. \end{align*}\vspace{-1em}

We refer to ${\bf L}_\text{sym}^k$ as the \textit{grouped decomposed components}, which can be used to measure information in a specific frequency band.

We now investigate the effect of random spatial augmentation on frequency bands through empirical study. Specifically, we aim to compare the spectrum difference of the graph before and after applying random spatial augmentation. We use  $\bf{L}_\text{sym}$ and $\bf{\tilde L}_\text{sym}$ to denote the Laplacian matrix in the original view and augmentation view, respectively. To measure their difference, a straightforward metric is   $F^k=\Vert{\bf L}_\text{sym}^k-{\bf{\tilde L}_\text{sym}}^k\Vert_F$ as adopted in~\cite{afgcl}. However, given that the decomposed components are scaled by the eigenvalues, we argue that they should be normalized by the corresponding eigenvalues for a fair comparison across different frequency bands. 
Thus, we normalize each grouped decomposed component ${\bf L}_\text{sym}^k$ by its largest eigenvalues $\operatorname{norm}({\bf L}_\text{sym}^k)$, and use 

\vspace{-1em}\begin{align*}F_{norm}^k=\Vert\frac{1}{\operatorname{norm}({\bf L}_\text{sym}^k)}{\bf L}_\text{sym}^k-\frac{1}{\operatorname{norm}({\bf{\tilde L}}_\text{sym}^k)}{\bf{\tilde L}_\text{sym}}^k\Vert_F, \end{align*}

\noindent where $\operatorname{\operatorname{norm}}({\bf L}_\text{sym}^k)={max(\lambda_i)},\ i\in[\frac{(k-1)N}{10},\frac{kN}{10})$, to quantity the spectrum change after augmentation. 

We take random edge insertion (injecting 20\% edges) as an example of random spatial augmentation, and then we check the $F_{norm}^k,\ k\in[1,10]$ between original and augmentation views. According to Fig.~\ref{fig:frob}, the impact of random spatial augmentation is approximately evenly distributed on all frequency components. However, from the perspective of the graph spectrum, the frequency carrying task-relevant information varies with the homophily degree: the low-frequency components are effective enough for homophilic graphs, while the heterophilic graphs take low-, middle- and high-frequency components~\cite{afgcl,RN248,RN249}. Falling short in treating frequency components with disparate considerations, such random spatial augmentation may be ineffective: when the informative and effective frequency bands are noticeably perturbed, the invariant information captured for the learned representations degenerates and becomes sub-optimal. In this work, we are motivated to propose graph augmentations with better invariant information from views of graph spectrum, in order to boost contrastive learning on graphs.

\section{The Proposed Framework\label{method}}
Our preliminary study in Section~\ref{spectraldifference} has shown that the impact of random spatial augmentation is approximately evenly distributed on all frequency components, and it is possible that task-relevant frequency components are noticeably corrupted in augmentation views. Therefore, this work aims to augment the graphs to preserve task-relevant frequency components and perturb the task-irrelevant ones with care. 

\begin{figure*}
\vspace{-1em}
\centering
\includegraphics[width=0.6\textwidth]{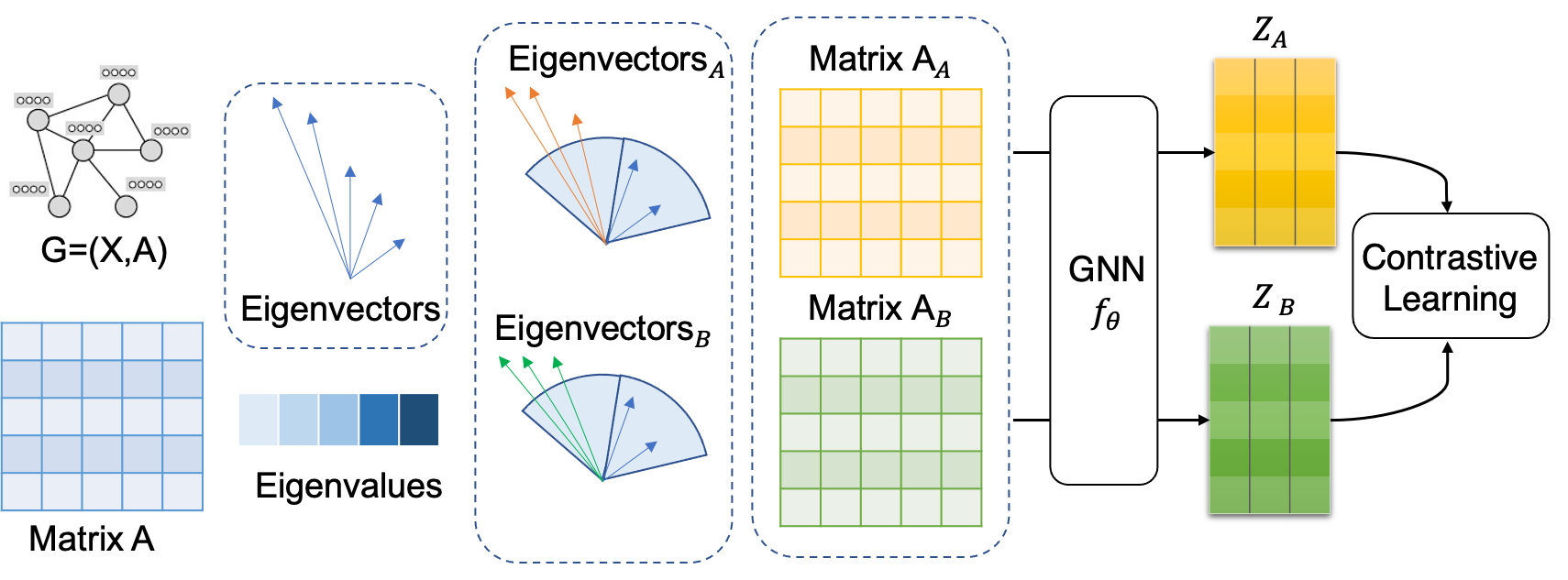}
\caption{An Overview of \ModelName.} 
\label{fig:framework}
\vspace{-1em}
\end{figure*}

To selectively augment the graph and avoid corrupting the task-relevant frequency components, we design a new contrastive learning framework \ModelName\ that augments the graph by considering the proper spectral impact on frequency bands. An overview of \ModelName{} is demonstrated in Figure~\ref{fig:framework}. As the key of \ModelName, the spectral augmentation consists of three main components: (i) selective perturbation on eigenpairs, (ii) reconstruction of adjacency matrix, and (iii) selective edge flipping. First, we inject perturbations to the selected frequency components while maintaining the invariant information; then based on the perturbed frequency components, we reconstruct the adjacency matrix for the augmentation view; and finally, guided by the new adjacency matrix with spectral augmentation, we select some edges to flip and obtain the structure augmentation of the original graph. Next, we will detail these three components {with the analysis of efficiency}. 

\vspace{-5pt}
\subsection{Selective Perturbation on Eigenpairs \label{selectband}}
Based on studies in Section~\ref{sec:pre} and previous works~\cite{RN248,RN249}, the augmentation of graphs should concentrate on distinct frequency bands w.r.t. homophily ratios. Therefore, our guiding strategy is to avoid corruption of task-relevant frequency components and to apportion perturbation to others. In this work, we perturb the frequency components by changing the directions of corresponding eigenvectors from $\{\bf\phi_i\}$ to $\{{\tilde{\bf{\phi}}}_i\}$, as they indicate the implicit properties of graph structure matrix\footnote{Unless stated otherwise, in Section~\ref{method} we set $A$ denoting the normalized adjacency matrix for simplicity.}. However, we face one major challenge: after perturbing the eigenvectors, how to maintain the orthogonality of eigenvectors and their spanned subspaces, consequently keeping the augmentation matrix qualified as a graph adjacency matrix? In this subsection, we aim to design proper perturbation and address this challenge simultaneously. 

As stated before, the frequency bands to be perturbed are expected to be task-irrelevant, and how to select them depends on the property (homophily ratio) of the graphs. Here we denote $\bB$ as the set of indexes of perturbed frequency components: given the eigen-decomposition of the adjacency matrix and the eigenpairs indexed with $\{0,1,\dots,n-1\}$, $\bB\subset\{0,1,\dots,n-1\}$ and $|\bB|\ll n$. We only inject the noises to the eigenpairs with indexes in $\bB$ to generate the perturbation for the adjacency matrix. For eigenvectors ${\bf \phi_i}\ (i\notin \bB)$, no perturbation is posed on them in order to keep the basic structure of the adjacency matrix: ${\tilde{\bf\phi}}_i={\bf \phi}_i,\ i\notin \bB$. In reality, we can determine the perturbed band $\bB$ selectively, by which the discrepancy of augmentation views is sufficient without corrupting the inherent semantics. For instance, when dealing with homophilic graphs where the low-frequency information is critical, we maintain the $b_0$ lowest frequency eigenvectors to preserve task-related information, and perturb the succeeding intermediate ones. On the other hand, for heterophilic graphs where the task-relevant information is spread across a broad spectrum of low, middle, and high-frequency bands, we introduce perturbations to the eigenvectors that are uniformly sampled from all these frequency bands.

Orthogonality among graph eigenvectors is an essential property that enables the reconstruction of the graph structure. This characteristic necessitates two conditions: first, the subspace spanned by the perturbed eigenvectors $\{{\tilde{\bf{\phi}}}_i\ |\ i\in \bB\}$ should be orthogonal to the unperturbed subspace $\{{{\bf{\phi}}}_i\ |\ i\notin \bB\}$. Second, the perturbed eigenvectors within the same subspace should be orthogonal to each other. 
To fulfill the first condition, we devise the perturbation as a linear combination of ${\bf \phi_i}\ (i\in \bB)$, as opposed to merely adding random noise. We generate the perturbed eigenvector ${\tilde{\bf{\phi}}}_i$ for eigenvectors ${{\bf{\phi}}}_i \ (i\in \bB)$ as follows:
\begin{align*}
{\tilde{\bf{\phi}}}_i=\sum\nolimits_{j\in \bB} \gamma_{ij}{{\bf{\phi}}}_j, \ i\in \bB,
\end{align*}\vspace{-1em}

\noindent where $\gamma_{ij}$ are random coefficients following a uniform distribution. To preserve the original semantics of the eigen-decomposition, we impose constraints such that $\sum_{j\in \bB}\gamma_{ij}=1$ and $\gamma_{ii}\gg \gamma_{ij}$.

To satisfy the second condition, we ensure that all the perturbed eigenvectors ${\tilde{\bf{\phi}}}_i$ in the subspace spanned by $\{{\tilde{\bf{\phi}}}_i\ |\ i\in \bB\}$ remain orthogonal to each other. Once a frequency band $\bB$ is perturbed, we retain the eigenvector corresponding to the lowest frequency in $\bB$. We then process the remaining ones sequentially using the Gram–Schmidt method, thus rendering $\tilde{\bf{\phi}}_i\ (i\in \bB)$ orthogonal to each other.
By doing so, the new eigenvector matrix upholds its orthogonality. This results in a new basis $\{\tilde{\bf{\phi}}\}$ that adheres to the properties of a graph structure matrix, enabling us to reconstruct a new matrix that qualifies as the adjacency matrix of an augmented graph.

\vspace{-5pt}
\subsection{Adjacency Matrix Reconstruction\label{matrixreconstruction}}
It is worth noting that most GNNs take the graph structure as input rather than the graph spectrum. Thus, we need to convert the aforementioned spectral perturbation to the spatial domain to obtain the augmented graph topology, which can be achieved by  reconstructing the adjacency matrix from the perturbed eigenvalues and eigenvecetors.  After injecting noises into eigenvectors for frequencies in $\bB$, one straightforward way to reconstruct the adjacency matrix is to add all the decomposed components together as below:  
\vspace{-1em}\begin{align} \label{reconstruct1}
{\tilde {\bf A}} = \sum\nolimits_{i=1}^n\omega_i{\tilde{\bf\phi}}_i{\tilde{\bf\phi}}_i^\top
= \sum\nolimits_{i\notin \bB}\omega_i{\bf \phi}_i{\bf \phi}_i^\top + \sum\nolimits_{i\in \bB}\omega_i{\tilde{\bf\phi}}_i{\tilde{\bf\phi}}_i^\top
\end{align}\vspace{-1em}

However, reconstructing the adjacency matrix as Eq.~\eqref{reconstruct1} requires all the eigenpairs. The  complexity of full eigen-decomposition is $\cO (n^3)$~\cite{pan1999complexity,span}, which is prohibitively expensive when the graph is large. In this work, we propose three strategies to reduce the computational complexity.

(i) As most of the frequency bands are not perturbed, we can reuse the frequency components $(i\notin \bB)$ and only consider the perturbed ones $(i\in \bB)$. Instead of reconstructing ${\tilde {\bf A}}$ from scratch, we adopt an incremental method by simply removing the selected components $\omega_i{\bf \phi}_i{\bf \phi}_i^\top,\ (i\in \bB)$ and add back the perturbed counterparts $\omega_i{\tilde{\bf\phi}}_i{\tilde{\bf\phi}}_i^\top,\ (i\in \bB)$. By rewriting Eq.~\eqref{reconstruct1} to Eq.~\eqref{reconstruct2a}, the complexity is reduced from $\cO (n^3)$ to $\cO (|\bB|\times n^2)$. 
\begin{equation} \label{reconstruct2a}
\begin{split}
{\tilde {\bf A}} =& {\bf A} - \sum\nolimits_{i\in \bB}\omega_i{\bf \phi}_i{\bf \phi}_i^\top + \sum\nolimits_{i\in \bB}\omega_i{\tilde{\bf\phi}}_i{\tilde{\bf\phi}}_i^\top\\
\end{split}
\end{equation}\vspace{-1em}

(ii) The second and third terms in Eq.~\eqref{reconstruct2a} can be merged together. Just as introduced in Section~\ref{selectband}, the perturbation is given by random coefficients $\gamma_{ij}$ while ${\tilde{\bf{\phi}}}$ can be given by the original ${\bf{\phi}}$. Keeping ${\tilde{\bf{\phi}}}$ in the format of combinations of ${\bf{\phi}}$, Eq.~\eqref{reconstruct2a} can be further simplified as Eq.~\eqref{reconstruct2b} with scalar operation of $\gamma_{ij}$ and $\omega_i$, where the cost of matrix operation is half of Eq.~\eqref{reconstruct2a}:

\vspace{-1em}
\begin{equation} \label{reconstruct2b}
\begin{split}
{\tilde {\bf A}}
=& {\bf A} - \sum\nolimits_{i\in \bB}\omega_i{\bf \phi}_i{\bf \phi}_i^\top + \sum\nolimits_{(i,j)\in \bB\times \bB} \omega_i\cdot \gamma_{ij}^2\cdot {\bf \phi}_j{\bf \phi}_j^\top\\
=& {\bf A} + \sum\nolimits_{i\in \bB}[(\sum\nolimits_{j\in \bB} \omega_j 
 \gamma_{ji}^2)-\omega_i]\cdot{\bf \phi}_i{\bf \phi}_i^\top
\end{split}
\end{equation} 

With the strategies above, the complexity is much alleviated and the computation of general graph data is feasible and acceptable. 

\vspace{-5pt}
\subsection{Selective Edge Flipping~\label{edgeflip}}
The graph structure perturbation method discussed above often results in dense matrices {with the perturbed dense eigenvectors}. These can significantly increase computational costs when used in GNNs. Consequently, it is crucial to sparsify these adjacency matrices to maintain computational efficiency.
Building upon the reconstructed adjacency matrix ${\tilde {\bf A}}$, we devise edge-flipping strategies for both existing and non-existing edges, focusing on their perturbation. As the edge flipping is contingent on the pairs of $({\tilde {\bf A}}_{ij},\ {\bf A}_{ij})$, it can be computationally challenging to calculate and store all the elements in the dense matrix ${\tilde {\bf A}}$ when dealing with large-scale graph data. In the following subsection, we elucidate the edge-flipping strategies for existing and non-existing edges separately and explore potential solutions to alleviate the computational complexity issue. 

For \textbf{existent edges}, we selectively \textbf{drop} some of them based on the relative perturbation in $\tilde {\bf A}$. Intuitively, the edges perturbed dramatically by the new basis $\{\bf\tilde\phi\}$ are less likely to be invariant among multiple views and should be dropped to provide contrastive supervision. For the preserved existent edges, the perturbation can also serve as changes to corresponding edge weights. 

By denoting $\Psi^{+}_{ij}=\Vert{\tilde {\bf A}}_{ij}-{\bf A}_{ij}\Vert/{\bf A}_{ij}$ as the relative perturbation for edge $(v_i,v_j)$, we sort the $\Psi^{+}_{ij}$ and remove the edges with $r_1\%$ largest $\Psi^{+}_{ij}$ values, where $r_1$ is the ratio of edge dropping. For the remaining edges, the values $\tilde {\bf A}_{ij}$ are regarded as the weights of edges re-scaled between 0 and 1, making the augmentation view a weighted graph.

To avoid calculating and storing the dense matrix ${\tilde {\bf A}}$, in the edge-dropping step we propose an individual formula to get the relative perturbation. Take an existent edge $(v_p,v_q),\ (v_p,v_q)\in\mathcal E$ as one example, $\tilde {\bf A}_{pq}$ in relative perturbation $\Psi^{+}_{ij}$ can be calculated individually following: 

\vspace{-1em}
\begin{equation}
\begin{aligned}\label{equ:relptb}
{\tilde {\bf A}}_{pq}=&{\bf A}_{pq}-\sum\nolimits_{i\in\bB}\omega_i{\bf \phi}_i[p]{\bf \phi}_i[q]+\sum\nolimits_{i\in \bB}\omega_i{\tilde{\bf\phi}}_i[p]{\tilde{\bf\phi}}_i[q]\\
=&{\bf A}_{pq}+\sum\nolimits_{i\in \bB}[(\sum\nolimits_{j\in \bB} \omega_j 
 \gamma_{ji}^2)-\omega_i]\cdot{{\bf\phi}}_i[p]{{\bf\phi}}_i[q]
\end{aligned}
\end{equation}
\vspace{-1em}

\noindent where ${\bf \phi}_i[p]$ is the $p_{th}$ element of ${\bf \phi}_i$. The complexity is further reduced to $\mathcal O(m|\bB|)$.

For \textbf{nonexistent edges} $(v_i,v_j)\notin\cE$ {potential to \textbf{add}}, our argument is that the edges eligible to be added must have positive and strong relative perturbation, which indicates the intense propensity to exist in augmentation views with new basis $\{\tilde{\bf{\phi}}\}$. 

We denote $\Psi^{-}_{ij}$ as the relative perturbation for nonexistent edges. Since there is no explicit reference as ${\bf A}_{ij}$ for the existent counterpart, we choose the degrees of nodes as substitutes for the propensity of linking nodes $(v_i,v_j)$. We denote the \textit{proxy adjacency matrix} as ${\bf A}^{\rho}$ where ${\bf A}^{\rho}_{ij} = \frac{1}{\sqrt{d_id_j}}$. Here the proxy adjacency matrix can be interpreted as a "\textit{what-if}" threshold, indicating "\textit{although there is no edge between $(v_i,v_j)$, what would the ${\bf A}_{ij}$ value be if this edge exists}". Then the relative perturbation is defined as $\Psi^{-}_{ij}={\tilde {\bf A}}_{ij}/{\bf A}^{\rho}_{ij}$, which can be regarded as the likelihood of adding a potential edge $(v_i,v_j)$, following which we add the corresponding edges whose corresponding $\Psi^{-}_{ij}$ are the largest $r_2\%$ ones, where $r_2$ is the ratio of edge-adding. Similar to edge dropping, the values ${\tilde {\bf A}}_{ij}$ are regarded as the weights of the added edges. 

To alleviate the complexity problem, we adopt the same strategy as shown by Eq.~\eqref{equ:relptb} in the edge-dropping step. In addition, as the number of nonexistent edges is much larger than that of existent edges\footnote{For example, the graph data Cora~\cite{gcn} has 2708 nodes and 5429 edges, then the number of nonexistent edges ($2708^2-5429$) is 1350 times larger than the number of existent edges}, we propose a sampling strategy on top of the individual formula. First we sample an edge pool $EP\subset\{(v_p,v_q)\ |\ (v_p,v_q)\notin\cE\}$ out of all the nonexistent edges; then we only calculate the corresponding $\Psi^{-}_{pq}$ and decide the edges to add within $EP$. With the proposed edge pool, the complexity is reduced to $\mathcal O(|EP||\bB|)$, where $|EP|$ is the number of nonexistent edges in the edge pool.

\vspace{-5pt}
\subsection{Efficiency Improvement}
{As an augmentation method concerning eigen-decomposition, the additional overhead compared with the previous random methods can not be ignored. Thanks to partial eigen-decomposition algorithms, the computational cost could be reduced especially when incorporated with the incremental strategies in Section~\ref{matrixreconstruction}. When the selected band $\bB$ is a successive set and near the border sides, we reduce the overhead by maintaining the $K$ lowest eigenvalues ($n\gg K=(b_0+|\bB|)$). Given the Lanczos Algorithm for selective eigen-decomposition, the complexity of eigen-decomposition will be reduced to $\mathcal O(K\cdot n^2)$\cite{lanczos}. Empirically, following the settings in Section~\ref{experiment}, the full decomposition takes 12.15s on PubMed ($n=19.717$) while the partial counterpart only takes 0.39s ($k=500$), which is in accordance with the theoretical expectations. The partial decomposition boosts the process and makes it feasible on large-scale graphs. Note that the decomposition is a one-time pre-computation process, so the overhead of Section~\ref{selectband}, \ref{matrixreconstruction} could be amortized to each augmentation step.}

{In addition, the time consumed by edge dropping in Section~\ref{edgeflip} of~\ModelName{} is 3.24s while the edge flipping takes 16.58s, which are totally acceptable for the one-time augmentation methods.}

\vspace{-5pt}
\subsection{Model Training}
After detailing our spectral augmentation method with selective spectral perturbation, we next introduce the proposed GCL framework \ModelName. Following works like \cite{cca,span,gca,hlcl}, \ModelName~first generates two augmentation views $\tilde{\mathcal G}_A=({\tilde {\bf X}}_A, {\tilde {\bf A}}_A)$ and $\tilde{\mathcal G}_B=({\tilde {\bf X}}_B, {\tilde {\bf A}}_B)$, {where ${\tilde {\bf X}}$ is augmented node features by random masking as convention}. Then we feed them into a shared encoder function $f_\theta$ to generate the node embeddings ${\bf Z}_{k\in\{A,B\}}=f_\theta({\tilde {\bf X}}_k, {\tilde {\bf A}}_k)$, where $\theta$ is the learnable parameters. The encoder is then optimized by a loss function $\mathcal L = g_\omega({\bf Z}_A, {\bf Z}_B)$ with hyper-parameters $\omega$. In the inference step, the encoder $f_\theta$ is frozen and embeddings are generated by ${\bf Z}=f_\theta({\bf X}, {\bf A})$. 

Following~\cite{cca}, in this work we adopt the standard GCN
models~\cite{gcn} as the encoder $f_\theta$. For the contrastive loss function $g_\omega$, we adopt the SSG-CCA feature-level objective:

\vspace{-1em}
\begin{equation}
\cL_{SSG-CCA}=\Vert\tilde{\mathbf{Z}}_A-\tilde{\mathbf{Z}}_B\Vert_F^2+\alpha(\Vert\tilde{\mathbf{Z}}_A^{\top} \tilde{\mathbf{Z}}_A-\mathbf{I}\Vert_F^2+\Vert\tilde{\mathbf{Z}}_B^{\top} \tilde{\mathbf{Z}}_B-\mathbf{I}\Vert_F^2)
\label{eq:loss}
\end{equation}
\vspace{-1em}

\noindent where $\tilde{\mathbf{Z}}=\frac{Z-\mu(Z)}{\sigma(Z)*\sqrt{n}}$ is the normalized node embeddings and $\alpha$ is a trade-off hyper-parameter. 

Although the proposed GCL framework is based on \cite{cca}, our spectral augmentation method is general that can be incorporated into many existing GCL frameworks and loss objectives. In Section~\ref{plugothers}, we empirically demonstrate that our augmentation method can advance various GCL frameworks. 

Furthermore, we provide the following theorem to show why the proposed framework can learn effective node representations: 

\vspace{-5pt}
\begin{theorem}\label{thm:3}
Suppose the label information $Y$ is only controlled by the high/low-frequency components and the classes are balanced. Denote the learned representation as $Z$. Then, minizing the first term in Eq.~\eqref{eq:loss}  is approximately maximizing the  mutual information between $I(Z, Y)$ while minimizing the entropy $H(Z)$.
\end{theorem}

\vspace{-5pt}
The proof can be found in Section~\ref{app:proof3}. The above theorem suggests that when using selective spectral augmentation, our framework can (1) promote the mutual information between the learned representation and downstream task, i.e., $I(Z, Y)$, and (2) diminish the uncertainty of learned representations, i.e., $H(Z)$. This demonstrates that the proposed method can indeed learn useful node representations that are helpful for predicting downstream tasks. 

\section{Theorem Proofs}
\setcounter{theorem}{0}

\vspace{-5pt}
\subsection{Proof of Theorem~\ref{thm:1}}

\begin{proof}\label{app:proof1}
Following the definition, the homophily $h_1$ for label vector $\mathbf{y}$ can be rewritten as:

\vspace{-1em}
\begin{align*}
    & h_1=\frac{1}{\sum\nolimits_{{v_i} \in \mathcal{V}} d_{v_i}} \sum\nolimits_{v_i \in \mathcal{V}}\left(d_{v_i}-\sum\nolimits_{{v_j} \in N(v_i)}\left(\mathbf{y}_{v_i}-\mathbf{y}_{v_j}\right)^2\right) \nonumber \\
    & =\frac{1}{\sum\nolimits_{{v_i} \in \mathcal{V}} d_{v_i}} \sum\nolimits_{{v_i} \in \mathcal{V}} d_{v_i}-\frac{1}{\sum\nolimits_{{v_i} \in \mathcal{V}} d_{v_i}} \sum\nolimits_{v_i \in \mathcal{V}} \sum\nolimits_{v_j \in N(v_i)}\left(\mathbf{y}_{v_i}-\mathbf{y}_{v_j}\right)^2\\ 
    & = 1 - \frac{1}{2m} \sum\nolimits_{v_i \in \mathcal{V}} \sum\nolimits_{v_j \in N(v_i)}\left(\mathbf{y}_{v_i}-\mathbf{y}_{v_j}\right)^2,  \nonumber
\end{align*}
\vspace{-1em}

\noindent where $m$ is the number of edges in $\mathcal{G}$. 

With $\mathbf{y}^{\top} \mathbf{L} \mathbf{y} = \sum_{v_i \in \mathcal{V}} \sum_{v_j \in N(v_i)}\left(\mathbf{y}_{v_i}-\mathbf{y}_{v_j}\right)^2 $, we have $h_1=1-\frac{1}{2m} \mathbf{y}^{\top} \mathbf{L} \mathbf{y}$. Similarly, $h_2=1-\frac{1}{2m} \mathbf{\hat{y}}^{\top} \mathbf{L} \mathbf{\hat{y}}$. Then we can obtain:

\vspace{-1em}
\begin{align*}
h_1-h_2 = -\frac{1}{2m} \mathbf{y}^{\top} \mathbf{L} \mathbf{y} +\frac{1}{2m} \mathbf{\hat{y}}^{\top} \mathbf{L} \mathbf{\hat{y}}.
\end{align*}
\vspace{-1em}

Furthermore, with $\mathbf{y}^{\top} \mathbf{L} \mathbf{y}=\left(\sum_i c_{i} \mathbf{u}^{\top}_i\right) \mathbf{L}\left(\sum_i c_i \mathbf{u}_i\right)=\sum_{i=0}^{n-1} c_i^2 \lambda_i \mathbf{u}^{\top}_i \mathbf{u}_i=\sum_{i=0}^{n-1} c_i^2 \lambda_i$, we can obtain:

\vspace{-1em}
\begin{equation*}
h_1 - h_2 =  -\frac{1}{2m} \sum\nolimits_{i=0}^{n-1} c_i^2 \lambda_i  + \frac{1}{2m} \sum\nolimits_{i=0}^{n-1} \hat{c}_i^2 \lambda_i.
\end{equation*}
\vspace{-1em}

Given $h_1 - h_2 = \Delta$, we have

\begin{equation}
\sum\nolimits_{i=0}^{n-1} \hat{c}_i^2 \lambda_i -\sum\nolimits_{i=0}^{n-1} c_i^2 \lambda_i = 2\Delta m. 
\label{eq:inequal}
\end{equation}
\vspace{-1em}

Then we proceed to prove that there exists an integer $M (0<M \leq n-1)$ such that $\sum_{i=M}^{n-1}\hat{c}^2_i \geq  \sum_{i=M}^{n-1}{c}^2_i + \frac{2\Delta m}{\lambda_M n} $. We prove this statement by contradiction. Suppose there does not exist an integer $M (0<M \leq n-1)$ that satisfies this condition. Thus, we have $\sum_{i=t}^{n-1}\hat{c}^2_i <  \sum_{i=t}^{n-1}{c}^2_i + \frac{2\Delta m}{\lambda_M n} $ {for} $M=1,\ldots, n-1$. Specifically, with $0=\lambda_0\leq\lambda_1\leq\lambda_2\leq\cdots\leq\lambda_{n-1}$, the following inequalities hold:

\vspace{-1em}
\begin{align*}
\lambda_0\left(\hat{c}_{0}^2+\hat{c}_{1}^2+\hat{c}_{2}^2+\cdots+\hat{c}_{n-1}^2\right) & =\lambda_0\left(c_{0}^2+c_{1}^2+{c}_{2}^2+\cdots+{c}_{n-1}^2  + \frac{2\Delta m}{\lambda_0 n}\right) \\
\left(\lambda_1-\lambda_0\right)\left(\hat{c}_{1}^2+\hat{c}_{2}^2+\cdots+\hat{c}_{n-1}^2\right) & \leq \left(\lambda_1-\lambda_0\right)\left({c}_{1}^2+\cdots+{c}_{n-1}^2+ \frac{2\Delta m}{\lambda_1 n}\right) \\
\left(\lambda_2-\lambda_1\right)\left(\hat{c}_{2}^2+\cdots+\hat{c}_{n-1}^2\right) & \leq 
\left(\lambda_2-\lambda_1\right)\left({c}_{2}^2+\cdots+{c}_{n-1}^2+ \frac{2\Delta m}{\lambda_2 n}\right) \\
& \cdots \\
\left(\lambda_{n-1}-\lambda_{n-2}\right) \hat{c}_{n-1}^2 & \leq \left(\lambda_{n-1}-\lambda_{n-2}\right) \left({c}_{n-1}^2 + \frac{2\Delta m}{\lambda_{|\mathcal{V}-1|}n}\right)
\end{align*}\nonumber
\vspace{-1em}

By summing over both sides of the above inequalities, we have

\vspace{-1em}
\begin{equation*}
\sum\nolimits_{i=0}^{n-1}\hat{c}^2_i \lambda_i \leq  \sum\nolimits_{i=0}^{n-1}{c}^2_i\lambda_i + 2{\Delta m} - \frac{2\Delta m}{n}\sum\nolimits_{i=1}^{n-1} \frac{\lambda_{i-1}}{\lambda_i}.
\end{equation*}

Plugging Eq.~\eqref{eq:inequal} into the above inequality, we have

\vspace{-1em}
\begin{equation*}
2 \Delta m \leq   2{\Delta m} - \frac{2\Delta m}{n}\sum\nolimits_{i=1}^{n-1} \frac{\lambda_{i-1}}{\lambda_i} ,
\end{equation*}

\noindent which contradicts the fact that $\sum_{i=1}^{n-1} \frac{\lambda_{i-1}}{\lambda_i} > 0$. Thus, the aforementioned assumption should not hold, and there must exist an integer $M (0<M \leq n-1)$ such that $\sum_{i=M}^{n-1}\hat{c}^2_i \geq  \sum_{i=M}^{n-1}{c}^2_i + \frac{2\Delta m}{\lambda_M n} $.

According to Parseval's identity, we have: 

\vspace{-1em}
\begin{equation*}
\label{eq:parseval}
  \|{\bf y}\|^2 = \sum\nolimits_{i=0}^{n-1} c_i^2, \quad  \|{\bf \hat{y}}\|^2 = \sum\nolimits_{i=0}^{n-1} \hat{c}_i^2.
\end{equation*}

As the classes are balanced, we have $\|{\bf y}\|^2=\|{\bf \hat{y}}\|^2$. Therefore, we know

\vspace{-1em}
\begin{equation*}
  \sum\nolimits_{i=0}^{M-1} c_i^2  + \sum\nolimits_{i=M}^{n-1} c_i^2 = \sum\nolimits_{i=0}^{n-1} c_i^2 =  \sum\nolimits_{i=0}^{n-1} \hat{c}_i^2 = \sum\nolimits_{i=0}^{M-1} \hat{c}_i^2  + \sum\nolimits_{i=M}^{n-1} \hat{c}_i^2  
\end{equation*}

Plugging this into  $\sum_{i=M}^{n-1}\hat{c}^2_i \geq  \sum_{i=M}^{n-1}{c}^2_i + \frac{2\Delta m}{\lambda_M n}$, we have 
$
    \sum_{i=0}^{M-1}\hat{c}^2_i\leq   \sum_{i=0}^{M-1}{c}^2_i - \frac{2\Delta m}{\lambda_M n},
$
which completes the proof.
\end{proof}

\vspace{-5pt}
\subsection{Proof of Theorem~\ref{thm:2}}
\begin{proof}\label{app:proof2}
We prove the theorem by contradiction. Suppose there does not exist an integer $M' (0<M' \leq n-1)$ that satisfies this condition. Thus, we have  $\sum_{i=M'}^{n-1} {c}^2_i \leq  \sum_{i=0}^{M'-1}{c}^2_i$ {for} $M'=1,\ldots, n-1$. Incorporating this inequality with the Parseval's identity, we have

\begin{equation*}
   \sum\nolimits_{i=M'}^{n-1} {c}^2_i \leq \frac{\|{\bf y}\|^2}{2} = \frac{n}{4},  \quad\text{for}\;  M'=1,\ldots, n-1 
\end{equation*}

With $0=\lambda_0\leq\lambda_1\leq\lambda_2\leq\cdots\leq\lambda_{n-1}$, the following inequalities hold:

\vspace{-1em}
\begin{equation*}
\begin{aligned}
\lambda_0\left(c_{0}^2+c_{1}^2+c_{2}^2+\cdots+c_{n-1}^2\right) & =\frac{n}{4}\lambda_0 \\
\left(\lambda_1-\lambda_0\right)\left(c_{1}^2+\cdots+c_{n-1}^2\right) & \leq \frac{n}{4}\left(\lambda_1-\lambda_0\right) \\
\left(\lambda_2-\lambda_1\right)\left(c_{2}^2+\cdots+c_{n-1}^2\right) & \leq 
\frac{n}{4} \left(\lambda_2-\lambda_1\right)\\
& \cdots \\
\left(\lambda_{n-1}-\lambda_{n-2}\right) c_{n-1}^2 & \leq \frac{n}{4}\left(\lambda_{n-1}-\lambda_{n-2}\right)
\end{aligned}\nonumber
\end{equation*}

By summing over both sides of the above inequalities, we have

\begin{equation*}
\sum\nolimits_{i=0}^{n-1}{c}^2_i \lambda_i \leq  \frac{n}{4} \lambda_{n-1}.
\end{equation*}

Since $h_1=1-\frac{1}{2m} \sum_{i=0}^{n-1}{c}^2_i \lambda_i$, we have $
   h_1 \geq  (1- \frac{n\lambda_{n-1}}{8m}) 
$,
which contradicts the condition that $h_1 < 1- \frac{n\lambda_\text{max}}{8m}$. Thus, the assumption should not hold, and there must exist an integer $M'(0<M' \leq n-1)$ such that $\sum_{i=M'}^{n-1} {c}^2_i <  \sum_{i=0}^{M'-1}{c}^2_i$, which completes the proof.

\textbf{Remark 1.} It is worth noting that from $h_1=1-\frac{1}{2m} \sum_{i=0}^{n-1}{c}^2_i \lambda_i$, we have $\sum_{i=0}^{n-1}{c}^2_i \lambda_i=0$ when $h_1=1$. This indicates $\lambda_0=\lambda_1=\ldots=\lambda_{|V|-1}=0$  or $c_0\neq c_1=c_2=\ldots=c_{|V|-1}=0$.
In this extreme case, the label vector ${\bf y}$ is proportional to the first eigenvector ${\bf u}_0$. 
\end{proof}
\vspace{-5pt}

\vspace{-5pt}
\subsection{Proof of Theorem~\ref{thm:3}}
\begin{proof}\label{app:proof3}
If the label information $Y$ is only controlled by the high-frequency (or low-frequency) components, our proposed augmentation only modifies the low-frequency (or high-frequency) components of the graph, which does not affect the true labels of the nodes while creating new data samples. We reformulate the first term of Eq.~\eqref{eq:loss} as $\sum^{n-1}_{i=0} \|{\bf z}_i - {\bf \hat{z}}_i\|^2$ with ${\bf z}_i$, ${\bf \hat{z}}_i$ being the $i$-th row of node embeddings ${\bf Z}_A$ and ${\bf Z}_B$, respectively. Note that the node embeddings are normalized and we have $\|{\bf z}_i\|=\|{\bf \hat{z}}_i\|=1$. Let $\stackrel{c}{=}$ denote equality up to a multiplicative and/or additive constant. Then we have:
\begin{align}
\label{eq:theorem3}
\sum\nolimits^{n-1}_{i=0} \|{\bf z}_i - {\bf \hat{z}}_i\|^2  = \sum\nolimits^{n-1}_{i=0}\left(2-2\cdot\hat{\bf z}^\top_i {\bf z}_i\right) \stackrel{c}{=} \sum\nolimits^{n-1}_{i=0}\left(-\hat{\bf z}^\top_i {\bf z}_i\right) 
\end{align}
\vspace{-1em}

According to Theorem 2 of~\cite{jin2023empowering}, when the augmentation generates a data view of the same class for the test nodes and the node classes are balanced, minimizing Eq.~\eqref{eq:theorem3} is approximately minimizing the class-conditional entropy $H(Z|Y)$ between features $Z$ and labels $Y$. Furthermore, since $H(Z|Y)=H(Z)-I(Z,Y)$, we know that minimizing the first term in Eq.~\eqref{eq:loss} is approximately maximizing $I(Z, Y)$ while minimizing $H(Z)$.
\end{proof}

\section{Experiments\label{experiment}}
In this section, we verify the effectiveness of \ModelName~ on both the homophilic and heterophilic graphs with extensive experiments. In particular, we aim to answer the following questions:
\begin{itemize}
    \item [Q1] Can the proposed \ModelName~improve the GCL performance on both homophilic and heterophilic graphs? (Section~\ref{mainresult})
    \item [Q2] Is the proposed \ModelName~flexible for more GCL frameworks? (Section~\ref{plugothers})
    \item [Q3] Can the proposed \ModelName~learn representations with better robustness? (Section~\ref{attackresult})
\end{itemize}

\vspace{-5pt}
\subsection{Experimental Settings}
\subsubsection{Datasets}
We perform experiments on 8 popular graph datasets of different homophily ratios. The homophilic datasets include Cora, CiteSeer, and PubMed~\cite{gcn}, Amazon-Computers and Amazon-Photo~\cite{RN257}. The heterophilic datasets are Chameleon, Squirrel~\cite{RN258}, and the Actor co-occurrence network~\cite{RN255}. These datasets represent several real-world scenarios, as well as different ratios of homophily. Details could be found in Table~\ref{table:dataset}. We adopt the public split settings  if exist; as for Amazon-Computers and Amazon-
Photo, we adopt the 1:1:8 training/validation/testing splits.
\begin{table}[h]
\vspace{- 10 pt}
\caption{Statistics of Datasets}
\centering
\vspace{- 10 pt}
\resizebox{0.45\textwidth}{!}{
\begin{tabular}{l|cccccc}
\toprule[1pt]
Dataset & \#Node & \#Edge & \#Class & \#Feature & $h$ \\ \hline
Cora & 2,708 & 5,429 & 7 & 1,433 & 0.81 \\
CiteSeer & 3,327 & 4,732 & 6 & 3,703  & 0.74 \\
PubMed & 19,717 & 44,338 & 3 & 500  & 0.80 \\
Am-Computers & 13,752 & 491,722 & 10 & 767  & 0.78 \\
Am-Photo & 7,650 & 238,162 & 8 & 745  & 0.83 \\ \hline
Chameleon & 2,277 & 36,101 & 5 & 2,325  & 0.23 \\
Squirrel & 5,201 & 217,073 & 5 & 2,089  & 0.22 \\
Actor & 7,600 & 33,544 & 5 & 931  & 0.22\\
\bottomrule[1pt]
\end{tabular}}
\label{table:dataset}
\vspace{- 15 pt}
\end{table}

\subsubsection{Baselines} 
We focus on the node classification task of graph data, especially in the self-supervised learning setting. Representative baseline models chosen as the comparison include: (1) the basic encoder models in the supervised setting, including MLP and GCN~\cite{gcn}; (2) self-supervised learning methods on graph neural networks. This group contains Deep Graph Infomax (DGI)~\cite{dgi}, Multi-view Graph Representation Learning (MVGRL)~\cite{mvgrl}, Graph Contrastive Representation Learning (GRACE)~\cite{grace}, and Bootstrapped Graph Latents (BGRL)~\cite{bgrl}; (3) state-of-the-art graph contrastive learning methods, especially those considering the design of augmentation. The representative works include Canonical Correlation Analysis inspired Self-Supervised Learning on Graphs (CCA-SSG)~\cite{cca}, Augmentation-Free GCL (AF-GCL)~\cite{afgcl}, Graph Contrastive learning with Adaptive augmentation (GCA)~\cite{gca}, and Spectral Augmentation in GCL (GCL-SPAN~\cite{span}).

\subsubsection{Evaluation Protocol}
We follow the linear evaluation scheme usually adopted for graph contrastive learning~\cite{dgi}. First, the encoder model is trained on all the nodes without supervision labels, guided by the augmentation views and contrastive losses. Then the encoder and its parameters are frozen and the embeddings of all the nodes are obtained from it. Finally, a linear classifier is trained with the labels of the training/validation set and tested by the testing set. The {accuracy of node classification} on the testing set is the metric for the quality of representations learned by the corresponding self-supervised model.

\subsubsection{Implementation Details of \ModelName}
For the proposed \ModelName, we choose the number of perturbed eigenvectors (i.e. $|\bB|$) from $\{100,200,500\}$; for the homophilic graphs, the skipped lowest frequency $b_0$ from $\{10,50,100\}$. The encoder $f_\theta$ is a standard GCN model~\cite{gcn} with one or two layers. For both the contrastive learning and the downstream tasks, Adam optimizer~\cite{kingma2014adam} is used for all experiments. Other hyper-parameters are chosen from the search space listed below; we select the hyper-parameters by grid search.

\begin{itemize}
    \item Pivot weight in linear combination $\gamma_{ii}$: $\{0.5,0.7,0.9\}$
    \item Trade-off hyper-parameter of $\cL_{SSG-CCA}$: $\{1e-4$, $5e-4$, $1e-3$, $5e-3$, $1e-2\}$
    \item Hidden units of encoder: $\{128, 256, 512\}$
    \item Edge flipping (adding and dropping) ratio: $\{0, 0.1, 0.2, 0.3, 0.4\}$
    \item Node feature masking ratio: $\{0, 0.1, 0.2, 0.3, 0.4\}$
    \item Dropout rate: $\{0, 0.3, 0.6\}$
\end{itemize}
\vspace{-.5em}

\vspace{-5pt}
\subsection{Performance Comparison\label{mainresult}}
To answer the first research question about the effectiveness of \ModelName~on both homophilic and heterophilic graphs, we compare the proposed method with representative models of {node classification}. The performance and the average rank of each algorithm are reported in Table~\ref{tablemain1}. In the first group of Table~\ref{tablemain1}, we show the supervised baselines MLP and GCN~\cite{gcn}; and the third group of this table reports the novel baselines concentrating on better augmentation for GCL, coining the selective perturbation methods on the edges of the graph.

As observed in Table~\ref{tablemain1}, the proposed spectral augmentation method achieves competitive or comparable performances on both the homophilic and the heterophilic datasets\footnote{If not specified, the results of baselines are quoted from~\cite{cca,mvgrl,grace,gca} following the public split; otherwise, we re-implement the source codes of authors.}. 
 Among all the self-supervised baselines, \ModelName\ achieves the best performance on 5 out of 8 datasets and is competitive (top 3) on 7 datasets. For example, on Cora, Am-Photo and Chameleon, our model outperforms the second best baseline by a margin larger than 1\%. While the target of GCA~\cite{gca} and GCL-SPAN~\cite{span} 
 is similar to ours, the \ModelName\ outperforms them on all the datasets. The superiority of the proposed model demonstrates the effectiveness of perturbing frequency components to generate augmentation views.
\begin{table*}[t]
\caption{Node classification performance on graph datasets. The metric is the mean and variance of accuracy (\%). \textit{Avg. Rank} stands for the average of rankings on 8 datasets, while \textit{Homo. Rank} and \textit{Hete. Rank} are the average ranking on 5 homophilic and 3 heterophilic datasets respectively. The best and top-3 results are highlighted with \textbf{bold} and \underline{underline}.}
\centering
\vspace{- 8 pt}
\resizebox{\textwidth}{!}{
\begin{threeparttable}
\begin{tabular}{l|ccccc|ccc|ccc}
\toprule[1pt]
 Dataset & Cora & CiteSeer & PubMed & Am-Comp & Am-Photo &   Chameleon & Squirrel & Actor & \begin{tabular}[c]{@{}c@{}}\textit{Avg.}\\ \textit{Rank}\end{tabular}& \begin{tabular}[c]{@{}c@{}}\textit{Homo.}\\ \textit{Rank}\end{tabular}& \begin{tabular}[c]{@{}c@{}}\textit{Hete.}\\ \textit{Rank}\end{tabular} \\ \hline
 MLP
 & 47.92\pn0.41 & 49.31\pn0.26 & 69.14\pn0.34 & 73.81\pn0.21 & 78.53\pn0.32 & 48.11\pn2.23 & 31.68\pn1.90 & 36.17\pn1.09 & / & / & / \\
 GCN\cite{gcn} & 81.54\pn0.68 & 70.73\pn0.65 & 79.16\pn0.25 & 86.51\pn0.54 & 92.42\pn0.22 & 67.96\pn1.82 & 54.47\pn1.17 & 30.31\pn0.98& /  & / & /\\ \hline
 DGI\cite{dgi} & 82.30\pn0.60 & 71.80\pn0.70 & 76.80\pn0.60 & 83.95\pn047 & 91.61\pn0.22 & 60.27\pn0.70 & 42.22\pn0.63 & 28.30\pn0.76 & 6.63 & 7.40 & 5.33\\
 MVGRL\cite{mvgrl} & {82.89\pn0.09} & \underline{73.30\pn0.50} & 79.96\pn0.05 & 87.52\pn0.11 & 91.74\pn0.07 & 53.81\pn1.09 & 38.75\pn1.32 & \textbf{32.09\pn1.07} & 5.38 & 5.00 & 6.00\\
 GRACE\cite{grace} & 81.90\pn0.40 & 71.20\pn0.50 & {78.72\pn0.13} & 86.25\pn0.25 & 92.15\pn0.24 & 52.04\pn0.4 & 39.22\pn0.48 & 28.27\pn0.43 & 7.75 & 7.60 & 	8.00\\
 BGRL\cite{bgrl} & 81.44\pn0.72 & 71.82\pn0.48 & 80.18\pn0.63 & \underline{89.62\pn0.37} & \underline{93.07\pn0.34} & \underline{64.86\pn0.63} & 46.24\pn0.70 & 28.80\pn0.54 & {4.38} & 4.60 & 	4.00\\ 
 CCA-SSG\cite{cca} & \underline{84.20\pn0.40} & \underline{73.10\pn0.30} & \underline{81.60\pn0.40} & \underline{88.74\pn0.28} & \underline{93.14\pn0.14} & 57.39\pn1.38 & 42.22\pn0.90 & 26.35\pn0.35 & \underline{3.88} & 	\underline{2.40} & 6.33\\
 AF-GCL\cite{afgcl} & 83.16\pn0.13 & {71.96\pn0.42} & 79.05\pn0.75 & \textbf{89.68\pn0.19} & 92.49\pn0.31 & \underline{65.28\pn0.53} & \textbf{52.10\pn0.67} & {28.94\pn0.69} & \underline{3.38} & \underline{4.00} & 	\underline{2.33}\\\hline
 GCA\cite{gca} & 82.07\pn0.10 & 71.33\pn0.37 & \underline{80.21\pn0.39} & 87.85\pn0.31 & 92.49\pn0.09 & 57.39\pn1.38 & 42.22\pn0.9 & 26.35\pn0.35 & 5.75 & 	5.40	 & 	6.33\\
 GCL-SPAN\cite{span}\tnote{$\dagger$} & \underline{83.30\pn0.49} & 71.75\pn0.49 & 79.27\pn0.15 & 80.89\pn0.15 & 91.24\pn0.23 & 62.94\pn0.66 & \underline{47.23\pn1.31} & \underline{29.82\pn0.10} & \underline{5.50}  & 6.80 & \underline{3.33}\\ \hline
 \textbf{\ModelName} & \textbf{85.27\pn0.10} & \textbf{75.41\pn0.84} & \textbf{83.00\pn0.61} & 88.67\pn0.15 & \textbf{93.17\pn0.31} & \textbf{67.23\pn1.78} & \underline{50.34\pn0.56} & \underline{31.47\pn0.04} & \textbf{1.63} & 	\textbf{1.60}	 & 	\textbf{1.67}\\
\bottomrule[1pt]
\end{tabular}
\begin{tablenotes}
\item[$\dagger$] Results are reproduced by us with the code of authors to follow the public splits.
\end{tablenotes}
\end{threeparttable}}
\label{tablemain1}
\vspace{-0.1in}
\end{table*}
\begin{table*}[thb]
\caption{Node classification performance on graph datasets. We plug the \ModelName{} augmentation to different GCL frameworks, denoted by suffix  \textit{+GASSER}. The better performances are highlighted in bold; the metric is accuracy (\%).}
\centering
\vspace{- 8 pt}
\resizebox{0.7\textwidth}{!}{
\begin{tabular}{l|ccc|ccc}
\toprule[1pt]
Dataset & Cora & CiteSeer & PubMed &  Chameleon & Squirrel & Actor \\\hline
GRACE\cite{grace} & 81.90\pn0.40 & 71.20\pn0.50 & 78.72\pn0.13 & 52.04\pn 0.45 & 39.22\pn 0.48 & 28.27\pn0.43 \\
\begin{tabular}[c]{@{}l@{}}GRACE\textit{+GASEER}\end{tabular} & \textbf{84.10\pn 0.26} & \textbf{74.47\pn0.64} & \textbf{83.97\pn 0.52} & \textbf{52.48\pn 0.13} & 36.89\pn 0.82 & \textbf{30.37\pn0.48} \\\hline
MVGRL\cite{mvgrl} & {82.89\pn0.09} & {73.30\pn0.50} & 79.96\pn0.05 & 53.81\pn1.09 & 38.75\pn1.32 & {32.09\pn1.07} \\
\begin{tabular}[c]{@{}l@{}}MVGRL\textit{+GASEER}\end{tabular} & 80.36\pn 0.05 & \textbf{74.48\pn0.73} & \textbf{80.80\pn 0.19} & \textbf{56.58\pn 0.41} & \textbf{40.79\pn 0.29} & \textbf{33.42\pn0.14} \\\hline
CCA-SSG\cite{cca} & {84.20\pn 0.40} & {73.10\pn 0.30} & {81.6\pn 0.40} & 57.39\pn 1.38 & 42.22\pn 0.90 & 26.35\pn 0.35  \\
\begin{tabular}[c]{@{}l@{}}CCA-SSG\textit{+GASEER}\end{tabular} & \textbf{85.27\pn 0.10} & \textbf{75.41\pn0.84} & \textbf{83.00\pn 0.61}  & \textbf{67.23\pn 1.78} & \textbf{50.34\pn 0.56} & \textbf{31.47\pn 0.04}\\
\bottomrule[1pt]
\end{tabular}}
\label{tablemain2}
\end{table*}
\begin{figure*}
\begin{center}
\includegraphics[width=0.75\textwidth]{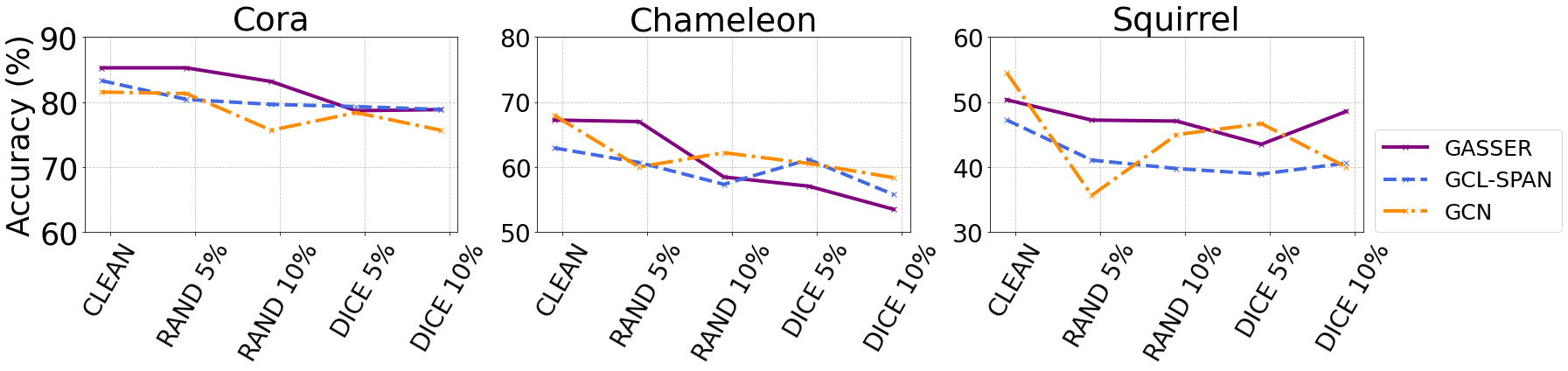}
\end{center}
\vspace{- 10 pt}
\caption{Node classification performance in attack setting on four datasets. \textit{RAND} stands for random attack strategies and \textit{DICE} stands for DICE~\cite{dice}; the following numbers indicate attack budgets.} 
\label{fig:att}
\vspace{-0.2in}
\end{figure*}

As one of the advantages of~\ModelName\ is fitting with both homophilic and heterophilic graphs, we show the performance rankings in Table~\ref{tablemain1}. We observe that~\ModelName\ achieves the best overall rankings of the two groups of graphs; although there are baselines that perform competitively on some datasets, they always lag behind~\ModelName\ on the other group of datasets. For example, CCA-SSG~\cite{cca} is the second best for homophilic graphs, but it works notably poorly with heterophilic graphs; AF-GCL~\cite{afgcl} is the contrasting example, which is competitive on heterophilic graphs but mediocre on homophilic graphs.

\vspace{-8.5pt}
\subsection{Integration of \ModelName\ into Existing GCL Frameworks\label{plugothers}}
In this experiment, we evaluate the effectiveness of \ModelName\ when combined with other GCL frameworks. The proposed model provides an augmentation-generating method, which can be easily incorporated into many GCL frameworks.  In particular, we replace the topology augmentation with the proposed spectrum-guided method in three graph contrastive learning frameworks for node classification, including GRACE~\cite{grace}, MVGRL~\cite{mvgrl}, and CCA-SSG~\cite{cca}. Note that GRACE adopts an instance-level objective, MVGRL uses the instance-graph objective, and CCA-SSG utilizes a feature-level objective. We choose these three representative contrastive objectives to demonstrate the broad integration of~\ModelName.

Table~\ref{tablemain2} shows our augmentation method advances the performance gains on all three contrastive learning frameworks. As we can see from this table, for each GCL framework, at least 5 out of the 6 representative datasets witness the performance gain with augmentation from \ModelName. This indicates that performance improvement is not dependent on the contrastive objective, no matter if it is local-local, local-global, or feature-level. In addition, all 3 heterophilic graph datasets achieve better results with \ModelName. While GCL on heterophilic graphs is understudied compared with the homophilic graphs, our proposed model can work well on them. Since \ModelName~perturbs the graph structure with care by keeping the important information and discarding the unimportant, the correspondence of augmentation views preserves much more task-relevant information than the random arbitrary methods, assisting GCL with more effective learning processes. 

\vspace{-8.5pt}
\subsection{Robustness of \ModelName \label{attackresult}}
We examine the robustness of \ModelName\ with the poisoned graphs. In practice, the augmentation and representations are derived from the original graphs, while the downstream tasks are trained and tested on structurally poisoned graphs. Here we choose two structure attack strategies, i.e., \textit{Random} and \textit{DICE}~\cite{dice}, and the perturbation budget $\sigma$ is selected from 0.05 and 0.1 (i.e. the number of flipped edges is $\sigma\times m$).

Figure~\ref{fig:att} shows performances in attack settings on four representative datasets. We  observe the encoders trained by \ModelName\ achieve a better balance between better robustness and higher performances. As \ModelName\ is trained with the graphs with corrupted structures and forced to capture the shared inherent semantics of input data, its encoders are more robust and invariant to the structure attacks.

Besides the experiments with the full settings, we also show the contribution of components of \ModelName. Fixing the overall framework and removing the edge weights in Section~\ref{edgeflip}, \ModelName-{W} is the variant with an unweighted graph. Specifically, after injecting noises with the spectral domain, we ignore the values as the weights of edges in augmentation views and assign the weights as 1 to all edges. By exchanging the selection of perturbed bands $\bB$ between homophilic and heterophilic graphs (Section~\ref{selectband}), \ModelName-{B} is the variant perturbing the spectrum on undesirable bands.

\begin{figure}
\begin{center}
\includegraphics[width=0.45\textwidth]{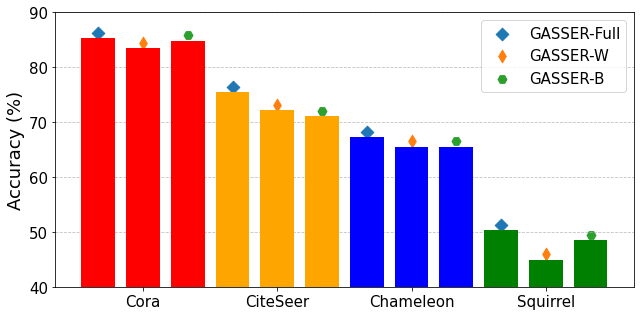}
\end{center}
\vspace{-1em}
\caption{Node classification performance of ablation study on four datasets. For each dataset, the results are of \ModelName-Full, \ModelName-W, and \ModelName-B respectively.} 
\label{fig:ablation}
\vspace{-15pt}
\end{figure}

Figure~\ref{fig:ablation} shows the comparison of the full model and its variants. The full version of \ModelName\ generally outperforms the two variants with margins, indicating the contributions of both the edge weights assigned by spectrum augmentation and the tailored frequency bands in \ModelName\ augmentation. 

\vspace{-5pt}
\section{Conclusions}

In this work, we propose a novel spectral augmentation method \ModelName~for graph contrastive learning, where the frequency bands are processed differently based on the properties of graphs. Beyond the existing works directly perturbing the graph structure in the spatial domain, \ModelName~perturbs eigenpairs of the adjacency matrix in the spectral domain and determines the edges to be perturbed 

\noindent based on the spectral hints. Extensive experiments of \ModelName~on homophilic and heterophilic graphs demonstrated the effectiveness, robustness, and generalizability of incorporating with other GCL frameworks. In the future, we plan to explore automatic methods to detect the most effective frequencies for graph augmentation.

\bibliographystyle{ACM-Reference-Format}
\bibliography{ref}


\begin{thebibliography}{62}


\ifx \showCODEN    \undefined \def \showCODEN     #1{\unskip}     \fi
\ifx \showDOI      \undefined \def \showDOI       #1{#1}\fi
\ifx \showISBNx    \undefined \def \showISBNx     #1{\unskip}     \fi
\ifx \showISBNxiii \undefined \def \showISBNxiii  #1{\unskip}     \fi
\ifx \showISSN     \undefined \def \showISSN      #1{\unskip}     \fi
\ifx \showLCCN     \undefined \def \showLCCN      #1{\unskip}     \fi
\ifx \shownote     \undefined \def \shownote      #1{#1}          \fi
\ifx \showarticletitle \undefined \def \showarticletitle #1{#1}   \fi
\ifx \showURL      \undefined \def \showURL       {\relax}        \fi
\providecommand\bibfield[2]{#2}
\providecommand\bibinfo[2]{#2}
\providecommand\natexlab[1]{#1}
\providecommand\showeprint[2][]{arXiv:#2}

\bibitem[Arora et~al\mbox{.}(2019)]%
        {arora2019theoretical}
\bibfield{author}{\bibinfo{person}{Sanjeev Arora}, \bibinfo{person}{Hrishikesh Khandeparkar}, \bibinfo{person}{Mikhail Khodak}, \bibinfo{person}{Orestis Plevrakis}, {and} \bibinfo{person}{Nikunj Saunshi}.} \bibinfo{year}{2019}\natexlab{}.
\newblock \showarticletitle{A theoretical analysis of contrastive unsupervised representation learning}. In \bibinfo{booktitle}{\emph{36th International Conference on Machine Learning, ICML 2019}}. International Machine Learning Society (IMLS), \bibinfo{pages}{9904--9923}.
\newblock


\bibitem[Battaglia et~al\mbox{.}(2018)]%
        {battaglia2018relational}
\bibfield{author}{\bibinfo{person}{Peter~W Battaglia}, \bibinfo{person}{Jessica~B Hamrick}, \bibinfo{person}{Victor Bapst}, \bibinfo{person}{Alvaro Sanchez-Gonzalez}, \bibinfo{person}{Vinicius Zambaldi}, \bibinfo{person}{Mateusz Malinowski}, \bibinfo{person}{Andrea Tacchetti}, \bibinfo{person}{David Raposo}, \bibinfo{person}{Adam Santoro}, \bibinfo{person}{Ryan Faulkner}, {et~al\mbox{.}}} \bibinfo{year}{2018}\natexlab{}.
\newblock \showarticletitle{Relational inductive biases, deep learning, and graph networks}.
\newblock \bibinfo{journal}{\emph{arXiv preprint arXiv:1806.01261}} (\bibinfo{year}{2018}).
\newblock


\bibitem[Bielak et~al\mbox{.}(2022)]%
        {bielak2022graph}
\bibfield{author}{\bibinfo{person}{Piotr Bielak}, \bibinfo{person}{Tomasz Kajdanowicz}, {and} \bibinfo{person}{Nitesh~V Chawla}.} \bibinfo{year}{2022}\natexlab{}.
\newblock \showarticletitle{Graph Barlow Twins: A self-supervised representation learning framework for graphs}.
\newblock \bibinfo{journal}{\emph{Knowledge-Based Systems}}  \bibinfo{volume}{256} (\bibinfo{year}{2022}), \bibinfo{pages}{109631}.
\newblock
\showISSN{0950-7051}


\bibitem[Bo et~al\mbox{.}(2021)]%
        {RN248}
\bibfield{author}{\bibinfo{person}{Deyu Bo}, \bibinfo{person}{Xiao Wang}, \bibinfo{person}{Chuan Shi}, {and} \bibinfo{person}{Huawei Shen}.} \bibinfo{year}{2021}\natexlab{}.
\newblock \showarticletitle{Beyond low-frequency information in graph convolutional networks}. In \bibinfo{booktitle}{\emph{Proceedings of the AAAI Conference on Artificial Intelligence}}, Vol.~\bibinfo{volume}{35}. \bibinfo{pages}{3950--3957}.
\newblock
\showISBNx{2374-3468}


\bibitem[Chang et~al\mbox{.}(2021)]%
        {chang2021notall}
\bibfield{author}{\bibinfo{person}{Heng Chang}, \bibinfo{person}{Yu Rong}, \bibinfo{person}{Tingyang Xu}, \bibinfo{person}{Yatao Bian}, \bibinfo{person}{Shiji Zhou}, \bibinfo{person}{Xin Wang}, \bibinfo{person}{Junzhou Huang}, {and} \bibinfo{person}{Wenwu Zhu}.} \bibinfo{year}{2021}\natexlab{}.
\newblock \showarticletitle{Not all low-pass filters are robust in graph convolutional networks}.
\newblock \bibinfo{journal}{\emph{Advances in Neural Information Processing Systems}}  \bibinfo{volume}{34} (\bibinfo{year}{2021}), \bibinfo{pages}{25058--25071}.
\newblock


\bibitem[Chen et~al\mbox{.}({[n.\,d.]})]%
        {RN134}
\bibfield{author}{\bibinfo{person}{Ting Chen}, \bibinfo{person}{Simon Kornblith}, \bibinfo{person}{Mohammad Norouzi}, {and} \bibinfo{person}{Geoffrey Hinton}.} \bibinfo{year}{[n.\,d.]}\natexlab{}.
\newblock \showarticletitle{A simple framework for contrastive learning of visual representations}. In \bibinfo{booktitle}{\emph{International conference on machine learning}}. \bibinfo{publisher}{PMLR}, \bibinfo{pages}{1597--1607}.
\newblock
\showISBNx{2640-3498}


\bibitem[Chi and Ma(2022)]%
        {chi2022enhancing}
\bibfield{author}{\bibinfo{person}{Hongliang Chi} {and} \bibinfo{person}{Yao Ma}.} \bibinfo{year}{2022}\natexlab{}.
\newblock \showarticletitle{Enhancing Graph Contrastive Learning with Node Similarity}.
\newblock \bibinfo{journal}{\emph{arXiv preprint arXiv:2208.06743}} (\bibinfo{year}{2022}).
\newblock


\bibitem[Chung(1997)]%
        {chung1997spectral}
\bibfield{author}{\bibinfo{person}{Fan~RK Chung}.} \bibinfo{year}{1997}\natexlab{}.
\newblock \bibinfo{booktitle}{\emph{Spectral graph theory}}. Vol.~\bibinfo{volume}{92}.
\newblock \bibinfo{publisher}{American Mathematical Soc.}
\newblock


\bibitem[Entezari et~al\mbox{.}(2020)]%
        {entezari2020all}
\bibfield{author}{\bibinfo{person}{Negin Entezari}, \bibinfo{person}{Saba~A Al-Sayouri}, \bibinfo{person}{Amirali Darvishzadeh}, {and} \bibinfo{person}{Evangelos~E Papalexakis}.} \bibinfo{year}{2020}\natexlab{}.
\newblock \showarticletitle{All you need is low (rank) defending against adversarial attacks on graphs}. In \bibinfo{booktitle}{\emph{Proceedings of the 13th International Conference on Web Search and Data Mining}}. \bibinfo{pages}{169--177}.
\newblock


\bibitem[Fan et~al\mbox{.}(2019a)]%
        {fan2019graph}
\bibfield{author}{\bibinfo{person}{Wenqi Fan}, \bibinfo{person}{Yao Ma}, \bibinfo{person}{Qing Li}, \bibinfo{person}{Yuan He}, \bibinfo{person}{Eric Zhao}, \bibinfo{person}{Jiliang Tang}, {and} \bibinfo{person}{Dawei Yin}.} \bibinfo{year}{2019}\natexlab{a}.
\newblock \showarticletitle{Graph neural networks for social recommendation}. In \bibinfo{booktitle}{\emph{The world wide web conference}}. \bibinfo{pages}{417--426}.
\newblock


\bibitem[Fan et~al\mbox{.}(2019b)]%
        {fan2019deep}
\bibfield{author}{\bibinfo{person}{Wenqi Fan}, \bibinfo{person}{Yao Ma}, \bibinfo{person}{Dawei Yin}, \bibinfo{person}{Jianping Wang}, \bibinfo{person}{Jiliang Tang}, {and} \bibinfo{person}{Qing Li}.} \bibinfo{year}{2019}\natexlab{b}.
\newblock \showarticletitle{Deep social collaborative filtering}. In \bibinfo{booktitle}{\emph{Proceedings of the 13th ACM Conference on Recommender Systems}}. \bibinfo{pages}{305--313}.
\newblock


\bibitem[Ghose et~al\mbox{.}(2023)]%
        {amur2023sagcl}
\bibfield{author}{\bibinfo{person}{Amur Ghose}, \bibinfo{person}{Yingxue Zhang}, \bibinfo{person}{Jianye Hao}, {and} \bibinfo{person}{Mark Coates}.} \bibinfo{year}{2023}\natexlab{}.
\newblock \showarticletitle{Spectral Augmentations for Graph Contrastive Learning}.
\newblock  (\bibinfo{year}{2023}), \bibinfo{pages}{11213--11266}.
\newblock


\bibitem[Hamilton et~al\mbox{.}(2017)]%
        {hamilton2017representation}
\bibfield{author}{\bibinfo{person}{William~L Hamilton}, \bibinfo{person}{Rex Ying}, {and} \bibinfo{person}{Jure Leskovec}.} \bibinfo{year}{2017}\natexlab{}.
\newblock \showarticletitle{Representation learning on graphs: Methods and applications}.
\newblock \bibinfo{journal}{\emph{arXiv preprint arXiv:1709.05584}} (\bibinfo{year}{2017}).
\newblock


\bibitem[Hassani and Khasahmadi({[n.\,d.]})]%
        {mvgrl}
\bibfield{author}{\bibinfo{person}{Kaveh Hassani} {and} \bibinfo{person}{Amir~Hosein Khasahmadi}.} \bibinfo{year}{[n.\,d.]}\natexlab{}.
\newblock \showarticletitle{Contrastive multi-view representation learning on graphs}. In \bibinfo{booktitle}{\emph{International Conference on Machine Learning}}. \bibinfo{publisher}{PMLR}, \bibinfo{pages}{4116--4126}.
\newblock
\showISBNx{2640-3498}


\bibitem[Jaiswal et~al\mbox{.}(2020)]%
        {RN254}
\bibfield{author}{\bibinfo{person}{Ashish Jaiswal}, \bibinfo{person}{Ashwin~Ramesh Babu}, \bibinfo{person}{Mohammad~Zaki Zadeh}, \bibinfo{person}{Debapriya Banerjee}, {and} \bibinfo{person}{Fillia Makedon}.} \bibinfo{year}{2020}\natexlab{}.
\newblock \showarticletitle{A survey on contrastive self-supervised learning}.
\newblock \bibinfo{journal}{\emph{Technologies}} \bibinfo{volume}{9}, \bibinfo{number}{1} (\bibinfo{year}{2020}), \bibinfo{pages}{2}.
\newblock
\showISSN{2227-7080}


\bibitem[Jin et~al\mbox{.}(2020)]%
        {jin2020graph}
\bibfield{author}{\bibinfo{person}{Wei Jin}, \bibinfo{person}{Yao Ma}, \bibinfo{person}{Xiaorui Liu}, \bibinfo{person}{Xianfeng Tang}, \bibinfo{person}{Suhang Wang}, {and} \bibinfo{person}{Jiliang Tang}.} \bibinfo{year}{2020}\natexlab{}.
\newblock \showarticletitle{Graph structure learning for robust graph neural networks}. In \bibinfo{booktitle}{\emph{Proceedings of the 26th ACM SIGKDD international conference on knowledge discovery \& data mining}}. \bibinfo{pages}{66--74}.
\newblock


\bibitem[Jin et~al\mbox{.}(2023)]%
        {jin2023empowering}
\bibfield{author}{\bibinfo{person}{Wei Jin}, \bibinfo{person}{Tong Zhao}, \bibinfo{person}{Jiayuan Ding}, \bibinfo{person}{Yozen Liu}, \bibinfo{person}{Jiliang Tang}, {and} \bibinfo{person}{Neil Shah}.} \bibinfo{year}{2023}\natexlab{}.
\newblock \showarticletitle{Empowering Graph Representation Learning with Test-Time Graph Transformation}. In \bibinfo{booktitle}{\emph{The Eleventh International Conference on Learning Representations}}.
\newblock
\urldef\tempurl%
\url{https://openreview.net/forum?id=Lnxl5pr018}
\showURL{%
\tempurl}


\bibitem[Kh¯c et~al\mbox{.}(2020)]%
        {grlsurvey}
\bibfield{author}{\bibinfo{person}{Phúc H.~Lê Kh¯c}, \bibinfo{person}{G. Healy}, {and} \bibinfo{person}{A. Smeaton}.} \bibinfo{year}{2020}\natexlab{}.
\newblock \showarticletitle{Contrastive Representation Learning: A Framework and Review}.
\newblock \bibinfo{journal}{\emph{IEEE Access}}  \bibinfo{volume}{8} (\bibinfo{year}{2020}), \bibinfo{pages}{193907--193934}.
\newblock


\bibitem[Kim et~al\mbox{.}(2022)]%
        {kim2022graph}
\bibfield{author}{\bibinfo{person}{Dongki Kim}, \bibinfo{person}{Jinheon Baek}, {and} \bibinfo{person}{Sung~Ju Hwang}.} \bibinfo{year}{2022}\natexlab{}.
\newblock \showarticletitle{Graph self-supervised learning with accurate discrepancy learning}.
\newblock \bibinfo{journal}{\emph{Advances in Neural Information Processing Systems}}  \bibinfo{volume}{35} (\bibinfo{year}{2022}), \bibinfo{pages}{14085--14098}.
\newblock


\bibitem[Kingma and Ba(2015)]%
        {kingma2014adam}
\bibfield{author}{\bibinfo{person}{Diederik~P. Kingma} {and} \bibinfo{person}{Jimmy Ba}.} \bibinfo{year}{2015}\natexlab{}.
\newblock \showarticletitle{Adam: {A} Method for Stochastic Optimization}. In \bibinfo{booktitle}{\emph{3rd International Conference on Learning Representations}}, \bibfield{editor}{\bibinfo{person}{Yoshua Bengio} {and} \bibinfo{person}{Yann LeCun}} (Eds.).
\newblock
\urldef\tempurl%
\url{http://arxiv.org/abs/1412.6980}
\showURL{%
\tempurl}


\bibitem[Kipf and Welling(2017)]%
        {gcn}
\bibfield{author}{\bibinfo{person}{Thomas~N. Kipf} {and} \bibinfo{person}{Max Welling}.} \bibinfo{year}{2017}\natexlab{}.
\newblock \showarticletitle{Semi-Supervised Classification with Graph Convolutional Networks}. In \bibinfo{booktitle}{\emph{International Conference on Learning Representations}}.
\newblock
\urldef\tempurl%
\url{https://openreview.net/forum?id=SJU4ayYgl}
\showURL{%
\tempurl}


\bibitem[Lee et~al\mbox{.}(2022)]%
        {lee2022augmentation}
\bibfield{author}{\bibinfo{person}{Namkyeong Lee}, \bibinfo{person}{Junseok Lee}, {and} \bibinfo{person}{Chanyoung Park}.} \bibinfo{year}{2022}\natexlab{}.
\newblock \showarticletitle{Augmentation-free self-supervised learning on graphs}. In \bibinfo{booktitle}{\emph{Proceedings of the AAAI Conference on Artificial Intelligence}}, Vol.~\bibinfo{volume}{36}. \bibinfo{pages}{7372--7380}.
\newblock


\bibitem[Li et~al\mbox{.}(2021)]%
        {li2021graph}
\bibfield{author}{\bibinfo{person}{Rui Li}, \bibinfo{person}{Xin Yuan}, \bibinfo{person}{Mohsen Radfar}, \bibinfo{person}{Peter Marendy}, \bibinfo{person}{Wei Ni}, \bibinfo{person}{Terence~J O'Brien}, {and} \bibinfo{person}{Pablo~M Casillas-Espinosa}.} \bibinfo{year}{2021}\natexlab{}.
\newblock \showarticletitle{Graph signal processing, graph neural network and graph learning on biological data: a systematic review}.
\newblock \bibinfo{journal}{\emph{IEEE Reviews in Biomedical Engineering}} (\bibinfo{year}{2021}).
\newblock


\bibitem[Li et~al\mbox{.}({[n.\,d.]})]%
        {RN249}
\bibfield{author}{\bibinfo{person}{Shouheng Li}, \bibinfo{person}{Dongwoo Kim}, {and} \bibinfo{person}{Qing Wang}.} \bibinfo{year}{[n.\,d.]}\natexlab{}.
\newblock \showarticletitle{Beyond low-pass filters: Adaptive feature propagation on graphs}. In \bibinfo{booktitle}{\emph{Machine Learning and Knowledge Discovery in Databases. Research Track: European Conference, ECML PKDD 2021, Bilbao, Spain, September 13–17, 2021, Proceedings, Part II 21}}. \bibinfo{publisher}{Springer}, \bibinfo{pages}{450--465}.
\newblock
\showISBNx{3030865193}


\bibitem[Lin et~al\mbox{.}(2022)]%
        {span}
\bibfield{author}{\bibinfo{person}{Lu Lin}, \bibinfo{person}{Jinghui Chen}, {and} \bibinfo{person}{Hongning Wang}.} \bibinfo{year}{2022}\natexlab{}.
\newblock \showarticletitle{Spectral Augmentation for Self-Supervised Learning on Graphs}. In \bibinfo{booktitle}{\emph{The Eleventh International Conference on Learning Representations}}.
\newblock


\bibitem[Liu et~al\mbox{.}(2022)]%
        {liu2022graph}
\bibfield{author}{\bibinfo{person}{Yixin Liu}, \bibinfo{person}{Ming Jin}, \bibinfo{person}{Shirui Pan}, \bibinfo{person}{Chuan Zhou}, \bibinfo{person}{Yu Zheng}, \bibinfo{person}{Feng Xia}, {and} \bibinfo{person}{S~Yu Philip}.} \bibinfo{year}{2022}\natexlab{}.
\newblock \showarticletitle{Graph self-supervised learning: A survey}.
\newblock \bibinfo{journal}{\emph{IEEE Transactions on Knowledge and Data Engineering}} \bibinfo{volume}{35}, \bibinfo{number}{6} (\bibinfo{year}{2022}), \bibinfo{pages}{5879--5900}.
\newblock


\bibitem[Pan and Chen(1999)]%
        {pan1999complexity}
\bibfield{author}{\bibinfo{person}{Victor~Y Pan} {and} \bibinfo{person}{Zhao~Q Chen}.} \bibinfo{year}{1999}\natexlab{}.
\newblock \showarticletitle{The complexity of the matrix eigenproblem}. In \bibinfo{booktitle}{\emph{Proceedings of the thirty-first annual ACM symposium on Theory of computing}}. \bibinfo{pages}{507--516}.
\newblock


\bibitem[Parlett and Scott(1979)]%
        {lanczos}
\bibfield{author}{\bibinfo{person}{Beresford~N Parlett} {and} \bibinfo{person}{David~S Scott}.} \bibinfo{year}{1979}\natexlab{}.
\newblock \showarticletitle{The Lanczos algorithm with selective orthogonalization}.
\newblock \bibinfo{journal}{\emph{Mathematics of computation}} \bibinfo{volume}{33}, \bibinfo{number}{145} (\bibinfo{year}{1979}), \bibinfo{pages}{217--238}.
\newblock
\showISSN{0025-5718}


\bibitem[Pei et~al\mbox{.}(2019)]%
        {RN255}
\bibfield{author}{\bibinfo{person}{Hongbin Pei}, \bibinfo{person}{Bingzhe Wei}, \bibinfo{person}{Kevin Chen-Chuan Chang}, \bibinfo{person}{Yu Lei}, {and} \bibinfo{person}{Bo Yang}.} \bibinfo{year}{2019}\natexlab{}.
\newblock \showarticletitle{Geom-GCN: Geometric Graph Convolutional Networks}. In \bibinfo{booktitle}{\emph{International Conference on Learning Representations}}.
\newblock


\bibitem[Perozzi et~al\mbox{.}(2014)]%
        {perozzi2014deepwalk}
\bibfield{author}{\bibinfo{person}{Bryan Perozzi}, \bibinfo{person}{Rami Al-Rfou}, {and} \bibinfo{person}{Steven Skiena}.} \bibinfo{year}{2014}\natexlab{}.
\newblock \showarticletitle{Deepwalk: Online learning of social representations}. In \bibinfo{booktitle}{\emph{Proceedings of the 20th ACM SIGKDD international conference on Knowledge discovery and data mining}}. \bibinfo{pages}{701--710}.
\newblock


\bibitem[Qiu et~al\mbox{.}({[n.\,d.]})]%
        {gcc}
\bibfield{author}{\bibinfo{person}{Jiezhong Qiu}, \bibinfo{person}{Qibin Chen}, \bibinfo{person}{Yuxiao Dong}, \bibinfo{person}{Jing Zhang}, \bibinfo{person}{Hongxia Yang}, \bibinfo{person}{Ming Ding}, \bibinfo{person}{Kuansan Wang}, {and} \bibinfo{person}{Jie Tang}.} \bibinfo{year}{[n.\,d.]}\natexlab{}.
\newblock \showarticletitle{GCC: Graph contrastive coding for graph neural network pre-training}. In \bibinfo{booktitle}{\emph{Proceedings of the 26th ACM SIGKDD International Conference on Knowledge Discovery \& Data Mining}}. \bibinfo{pages}{1150--1160}.
\newblock


\bibitem[Robinson et~al\mbox{.}(2020)]%
        {robinson2020contrastive}
\bibfield{author}{\bibinfo{person}{Joshua~David Robinson}, \bibinfo{person}{Ching-Yao Chuang}, \bibinfo{person}{Suvrit Sra}, {and} \bibinfo{person}{Stefanie Jegelka}.} \bibinfo{year}{2020}\natexlab{}.
\newblock \showarticletitle{Contrastive Learning with Hard Negative Samples}. In \bibinfo{booktitle}{\emph{International Conference on Learning Representations}}.
\newblock


\bibitem[Rozemberczki et~al\mbox{.}(2021)]%
        {RN258}
\bibfield{author}{\bibinfo{person}{Benedek Rozemberczki}, \bibinfo{person}{Carl Allen}, {and} \bibinfo{person}{Rik Sarkar}.} \bibinfo{year}{2021}\natexlab{}.
\newblock \showarticletitle{Multi-scale attributed node embedding}.
\newblock \bibinfo{journal}{\emph{Journal of Complex Networks}} \bibinfo{volume}{9}, \bibinfo{number}{2} (\bibinfo{year}{2021}), \bibinfo{pages}{cnab014}.
\newblock
\showISSN{2051-1310}


\bibitem[Shchur et~al\mbox{.}(2018)]%
        {RN257}
\bibfield{author}{\bibinfo{person}{Oleksandr Shchur}, \bibinfo{person}{Maximilian Mumme}, \bibinfo{person}{Aleksandar Bojchevski}, {and} \bibinfo{person}{Stephan Günnemann}.} \bibinfo{year}{2018}\natexlab{}.
\newblock \showarticletitle{Pitfalls of graph neural network evaluation}.
\newblock \bibinfo{journal}{\emph{arXiv preprint arXiv:1811.05868}} (\bibinfo{year}{2018}).
\newblock


\bibitem[Spielman(2019)]%
        {sagt}
\bibfield{author}{\bibinfo{person}{Daniel Spielman}.} \bibinfo{year}{2019}\natexlab{}.
\newblock \bibinfo{booktitle}{\emph{Spectral and Algebraic Graph Theory (Incomplete Draft)}}. \bibinfo{series}{Yale lecture notes, draft of December}, Vol.~\bibinfo{volume}{4}.
\newblock 47 pages.
\newblock


\bibitem[Sun et~al\mbox{.}(2023)]%
        {sun2023graph}
\bibfield{author}{\bibinfo{person}{Dengdi Sun}, \bibinfo{person}{Mingxin Cao}, \bibinfo{person}{Zhuanlian Ding}, {and} \bibinfo{person}{Bin Luo}.} \bibinfo{year}{2023}\natexlab{}.
\newblock \showarticletitle{Graph Contrastive Learning with Intrinsic Augmentations}. In \bibinfo{booktitle}{\emph{Bio-Inspired Computing: Theories and Applications: 17th International Conference, BIC-TA 2022, Wuhan, China, December 16--18, 2022, Revised Selected Papers}}. Springer, \bibinfo{pages}{343--357}.
\newblock


\bibitem[Sun et~al\mbox{.}(2019)]%
        {RN202}
\bibfield{author}{\bibinfo{person}{Fan-Yun Sun}, \bibinfo{person}{Jordan Hoffman}, \bibinfo{person}{Vikas Verma}, {and} \bibinfo{person}{Jian Tang}.} \bibinfo{year}{2019}\natexlab{}.
\newblock \showarticletitle{InfoGraph: Unsupervised and Semi-supervised Graph-Level Representation Learning via Mutual Information Maximization}. In \bibinfo{booktitle}{\emph{International Conference on Learning Representations}}.
\newblock


\bibitem[Suresh et~al\mbox{.}(2021)]%
        {suresh2021adversarial}
\bibfield{author}{\bibinfo{person}{Susheel Suresh}, \bibinfo{person}{Pan Li}, \bibinfo{person}{Cong Hao}, {and} \bibinfo{person}{Jennifer Neville}.} \bibinfo{year}{2021}\natexlab{}.
\newblock \showarticletitle{Adversarial graph augmentation to improve graph contrastive learning}.
\newblock \bibinfo{journal}{\emph{Advances in Neural Information Processing Systems}}  \bibinfo{volume}{34} (\bibinfo{year}{2021}), \bibinfo{pages}{15920--15933}.
\newblock


\bibitem[Thakoor et~al\mbox{.}({[n.\,d.]})]%
        {bgrl}
\bibfield{author}{\bibinfo{person}{Shantanu Thakoor}, \bibinfo{person}{Corentin Tallec}, \bibinfo{person}{Mohammad~Gheshlaghi Azar}, \bibinfo{person}{Rémi Munos}, \bibinfo{person}{Petar Veličković}, {and} \bibinfo{person}{Michal Valko}.} \bibinfo{year}{[n.\,d.]}\natexlab{}.
\newblock \showarticletitle{Bootstrapped representation learning on graphs}. In \bibinfo{booktitle}{\emph{ICLR 2021 Workshop on Geometrical and Topological Representation Learning}}.
\newblock


\bibitem[Tian et~al\mbox{.}(2020)]%
        {tian2020makes}
\bibfield{author}{\bibinfo{person}{Yonglong Tian}, \bibinfo{person}{Chen Sun}, \bibinfo{person}{Ben Poole}, \bibinfo{person}{Dilip Krishnan}, \bibinfo{person}{Cordelia Schmid}, {and} \bibinfo{person}{Phillip Isola}.} \bibinfo{year}{2020}\natexlab{}.
\newblock \showarticletitle{What makes for good views for contrastive learning?}
\newblock \bibinfo{journal}{\emph{Advances in neural information processing systems}}  \bibinfo{volume}{33} (\bibinfo{year}{2020}), \bibinfo{pages}{6827--6839}.
\newblock


\bibitem[Tong et~al\mbox{.}(2006)]%
        {tong2006fast}
\bibfield{author}{\bibinfo{person}{Hanghang Tong}, \bibinfo{person}{Christos Faloutsos}, {and} \bibinfo{person}{Jia-Yu Pan}.} \bibinfo{year}{2006}\natexlab{}.
\newblock \showarticletitle{Fast random walk with restart and its applications}. In \bibinfo{booktitle}{\emph{Sixth international conference on data mining (ICDM'06)}}. IEEE, \bibinfo{pages}{613--622}.
\newblock


\bibitem[Trivedi et~al\mbox{.}(2022)]%
        {trivedi2022augmentations}
\bibfield{author}{\bibinfo{person}{Puja Trivedi}, \bibinfo{person}{Ekdeep~Singh Lubana}, \bibinfo{person}{Yujun Yan}, \bibinfo{person}{Yaoqing Yang}, {and} \bibinfo{person}{Danai Koutra}.} \bibinfo{year}{2022}\natexlab{}.
\newblock \showarticletitle{Augmentations in graph contrastive learning: Current methodological flaws \& towards better practices}. In \bibinfo{booktitle}{\emph{Proceedings of the ACM Web Conference 2022}}. \bibinfo{pages}{1538--1549}.
\newblock


\bibitem[Tsai et~al\mbox{.}(2020)]%
        {tsai2020self}
\bibfield{author}{\bibinfo{person}{Yao-Hung~Hubert Tsai}, \bibinfo{person}{Yue Wu}, \bibinfo{person}{Ruslan Salakhutdinov}, {and} \bibinfo{person}{Louis-Philippe Morency}.} \bibinfo{year}{2020}\natexlab{}.
\newblock \showarticletitle{Self-supervised Learning from a Multi-view Perspective}. In \bibinfo{booktitle}{\emph{International Conference on Learning Representations}}.
\newblock


\bibitem[Veli{\v{c}}kovi{\'c} et~al\mbox{.}(2018)]%
        {dgi}
\bibfield{author}{\bibinfo{person}{Petar Veli{\v{c}}kovi{\'c}}, \bibinfo{person}{William Fedus}, \bibinfo{person}{William~L Hamilton}, \bibinfo{person}{Pietro Li{\`o}}, \bibinfo{person}{Yoshua Bengio}, {and} \bibinfo{person}{R~Devon Hjelm}.} \bibinfo{year}{2018}\natexlab{}.
\newblock \showarticletitle{Deep Graph Infomax}. In \bibinfo{booktitle}{\emph{International Conference on Learning Representations}}.
\newblock


\bibitem[Wang et~al\mbox{.}(2022a)]%
        {afgcl}
\bibfield{author}{\bibinfo{person}{Haonan Wang}, \bibinfo{person}{Jieyu Zhang}, \bibinfo{person}{Qi Zhu}, {and} \bibinfo{person}{Wei Huang}.} \bibinfo{year}{2022}\natexlab{a}.
\newblock \showarticletitle{Augmentation-Free Graph Contrastive Learning with Performance Guarantee}.
\newblock \bibinfo{journal}{\emph{arXiv preprint arXiv:2204.04874}} (\bibinfo{year}{2022}).
\newblock


\bibitem[Wang et~al\mbox{.}(2022b)]%
        {wang2022can}
\bibfield{author}{\bibinfo{person}{Haonan Wang}, \bibinfo{person}{Jieyu Zhang}, \bibinfo{person}{Qi Zhu}, {and} \bibinfo{person}{Wei Huang}.} \bibinfo{year}{2022}\natexlab{b}.
\newblock \showarticletitle{Can Single-Pass Contrastive Learning Work for Both Homophilic and Heterophilic Graph?}
\newblock \bibinfo{journal}{\emph{arXiv preprint arXiv:2211.10890}} (\bibinfo{year}{2022}).
\newblock


\bibitem[Wang et~al\mbox{.}(2021)]%
        {wang2021review}
\bibfield{author}{\bibinfo{person}{Jianian Wang}, \bibinfo{person}{Sheng Zhang}, \bibinfo{person}{Yanghua Xiao}, {and} \bibinfo{person}{Rui Song}.} \bibinfo{year}{2021}\natexlab{}.
\newblock \showarticletitle{A review on graph neural network methods in financial applications}.
\newblock \bibinfo{journal}{\emph{arXiv preprint arXiv:2111.15367}} (\bibinfo{year}{2021}).
\newblock


\bibitem[Wang and Isola(2020)]%
        {wang2020understanding}
\bibfield{author}{\bibinfo{person}{Tongzhou Wang} {and} \bibinfo{person}{Phillip Isola}.} \bibinfo{year}{2020}\natexlab{}.
\newblock \showarticletitle{Understanding contrastive representation learning through alignment and uniformity on the hypersphere}. In \bibinfo{booktitle}{\emph{International Conference on Machine Learning}}. PMLR, \bibinfo{pages}{9929--9939}.
\newblock


\bibitem[Waniek et~al\mbox{.}(2018)]%
        {dice}
\bibfield{author}{\bibinfo{person}{Marcin Waniek}, \bibinfo{person}{Tomasz~P Michalak}, \bibinfo{person}{Michael~J Wooldridge}, {and} \bibinfo{person}{Talal Rahwan}.} \bibinfo{year}{2018}\natexlab{}.
\newblock \showarticletitle{Hiding individuals and communities in a social network}.
\newblock \bibinfo{journal}{\emph{Nature Human Behaviour}} \bibinfo{volume}{2}, \bibinfo{number}{2} (\bibinfo{year}{2018}), \bibinfo{pages}{139--147}.
\newblock


\bibitem[Wu et~al\mbox{.}(2020)]%
        {RN207}
\bibfield{author}{\bibinfo{person}{Mike Wu}, \bibinfo{person}{Milan Mosse}, \bibinfo{person}{Chengxu Zhuang}, \bibinfo{person}{Daniel Yamins}, {and} \bibinfo{person}{Noah Goodman}.} \bibinfo{year}{2020}\natexlab{}.
\newblock \showarticletitle{Conditional Negative Sampling for Contrastive Learning of Visual Representations}.
\newblock  (\bibinfo{year}{2020}).
\newblock


\bibitem[Xu et~al\mbox{.}(2018)]%
        {xu2018powerful}
\bibfield{author}{\bibinfo{person}{Keyulu Xu}, \bibinfo{person}{Weihua Hu}, \bibinfo{person}{Jure Leskovec}, {and} \bibinfo{person}{Stefanie Jegelka}.} \bibinfo{year}{2018}\natexlab{}.
\newblock \showarticletitle{How Powerful are Graph Neural Networks?}. In \bibinfo{booktitle}{\emph{International Conference on Learning Representations}}.
\newblock


\bibitem[Yang and Mirzasoleiman(2023)]%
        {hlcl}
\bibfield{author}{\bibinfo{person}{Wenhan Yang} {and} \bibinfo{person}{Baharan Mirzasoleiman}.} \bibinfo{year}{2023}\natexlab{}.
\newblock \showarticletitle{Graph Contrastive Learning Under Heterophily: Utilizing Graph Filters to Generate Graph Views}.
\newblock \bibinfo{journal}{\emph{Submitted to The Eleventh International Conference on Learning Representations}} (\bibinfo{year}{2023}).
\newblock
\urldef\tempurl%
\url{https://openreview.net/forum?id=NzcUQuhEGef}
\showURL{%
\tempurl}


\bibitem[Ye et~al\mbox{.}(2020)]%
        {ye2020symmetrical}
\bibfield{author}{\bibinfo{person}{Shuqian Ye}, \bibinfo{person}{Jiechun Liang}, \bibinfo{person}{Rulin Liu}, {and} \bibinfo{person}{Xi Zhu}.} \bibinfo{year}{2020}\natexlab{}.
\newblock \showarticletitle{Symmetrical graph neural network for quantum chemistry with dual real and momenta space}.
\newblock \bibinfo{journal}{\emph{The Journal of Physical Chemistry A}} \bibinfo{volume}{124}, \bibinfo{number}{34} (\bibinfo{year}{2020}), \bibinfo{pages}{6945--6953}.
\newblock


\bibitem[You et~al\mbox{.}(2020)]%
        {RN55}
\bibfield{author}{\bibinfo{person}{Yuning You}, \bibinfo{person}{Tianlong Chen}, \bibinfo{person}{Yongduo Sui}, \bibinfo{person}{Ting Chen}, \bibinfo{person}{Zhangyang Wang}, {and} \bibinfo{person}{Yang Shen}.} \bibinfo{year}{2020}\natexlab{}.
\newblock \showarticletitle{Graph contrastive learning with augmentations}.
\newblock \bibinfo{journal}{\emph{Advances in Neural Information Processing Systems}}  \bibinfo{volume}{33} (\bibinfo{year}{2020}), \bibinfo{pages}{5812--5823}.
\newblock


\bibitem[Zhang et~al\mbox{.}(2021)]%
        {cca}
\bibfield{author}{\bibinfo{person}{Hengrui Zhang}, \bibinfo{person}{Qitian Wu}, \bibinfo{person}{Junchi Yan}, \bibinfo{person}{David Wipf}, {and} \bibinfo{person}{Philip~S Yu}.} \bibinfo{year}{2021}\natexlab{}.
\newblock \showarticletitle{From canonical correlation analysis to self-supervised graph neural networks}.
\newblock \bibinfo{journal}{\emph{Advances in Neural Information Processing Systems}}  \bibinfo{volume}{34} (\bibinfo{year}{2021}), \bibinfo{pages}{76--89}.
\newblock


\bibitem[Zhang et~al\mbox{.}(2022)]%
        {zhang2022costa}
\bibfield{author}{\bibinfo{person}{Yifei Zhang}, \bibinfo{person}{Hao Zhu}, \bibinfo{person}{Zixing Song}, \bibinfo{person}{Piotr Koniusz}, {and} \bibinfo{person}{Irwin King}.} \bibinfo{year}{2022}\natexlab{}.
\newblock \showarticletitle{COSTA: Covariance-Preserving Feature Augmentation for Graph Contrastive Learning}. In \bibinfo{booktitle}{\emph{Proceedings of the 28th ACM SIGKDD Conference on Knowledge Discovery and Data Mining}}. \bibinfo{pages}{2524--2534}.
\newblock


\bibitem[Zhang et~al\mbox{.}(2023)]%
        {sfa}
\bibfield{author}{\bibinfo{person}{Yifei Zhang}, \bibinfo{person}{Hao Zhu}, \bibinfo{person}{Zixing Song}, \bibinfo{person}{Piotr Koniusz}, {and} \bibinfo{person}{Irwin King}.} \bibinfo{year}{2023}\natexlab{}.
\newblock \showarticletitle{Spectral feature augmentation for graph contrastive learning and beyond}. In \bibinfo{booktitle}{\emph{Proceedings of the AAAI Conference on Artificial Intelligence}}, Vol.~\bibinfo{volume}{37}. \bibinfo{pages}{11289--11297}.
\newblock


\bibitem[Zhao et~al\mbox{.}({[n.\,d.]})]%
        {auggraph}
\bibfield{author}{\bibinfo{person}{Tong Zhao}, \bibinfo{person}{Yozen Liu}, \bibinfo{person}{Leonardo Neves}, \bibinfo{person}{Oliver Woodford}, \bibinfo{person}{Meng Jiang}, {and} \bibinfo{person}{Neil Shah}.} \bibinfo{year}{[n.\,d.]}\natexlab{}.
\newblock \showarticletitle{Data augmentation for graph neural networks}. In \bibinfo{booktitle}{\emph{Proceedings of the aaai conference on artificial intelligence}}, Vol.~\bibinfo{volume}{35}. \bibinfo{pages}{11015--11023}.
\newblock
\showISBNx{2374-3468}


\bibitem[Zhou et~al\mbox{.}(2020)]%
        {zhou2020graph}
\bibfield{author}{\bibinfo{person}{Jie Zhou}, \bibinfo{person}{Ganqu Cui}, \bibinfo{person}{Shengding Hu}, \bibinfo{person}{Zhengyan Zhang}, \bibinfo{person}{Cheng Yang}, \bibinfo{person}{Zhiyuan Liu}, \bibinfo{person}{Lifeng Wang}, \bibinfo{person}{Changcheng Li}, {and} \bibinfo{person}{Maosong Sun}.} \bibinfo{year}{2020}\natexlab{}.
\newblock \showarticletitle{Graph neural networks: A review of methods and applications}.
\newblock \bibinfo{journal}{\emph{AI open}}  \bibinfo{volume}{1} (\bibinfo{year}{2020}), \bibinfo{pages}{57--81}.
\newblock


\bibitem[Zhu et~al\mbox{.}(2020b)]%
        {zhu2020beyond}
\bibfield{author}{\bibinfo{person}{Jiong Zhu}, \bibinfo{person}{Yujun Yan}, \bibinfo{person}{Lingxiao Zhao}, \bibinfo{person}{Mark Heimann}, \bibinfo{person}{Leman Akoglu}, {and} \bibinfo{person}{Danai Koutra}.} \bibinfo{year}{2020}\natexlab{b}.
\newblock \showarticletitle{Beyond homophily in graph neural networks: Current limitations and effective designs}.
\newblock \bibinfo{journal}{\emph{Advances in Neural Information Processing Systems}}  \bibinfo{volume}{33} (\bibinfo{year}{2020}), \bibinfo{pages}{7793--7804}.
\newblock


\bibitem[Zhu et~al\mbox{.}({[n.\,d.]})]%
        {gca}
\bibfield{author}{\bibinfo{person}{Yanqiao Zhu}, \bibinfo{person}{Yichen Xu}, \bibinfo{person}{Feng Yu}, \bibinfo{person}{Qiang Liu}, \bibinfo{person}{Shu Wu}, {and} \bibinfo{person}{Liang Wang}.} \bibinfo{year}{[n.\,d.]}\natexlab{}.
\newblock \showarticletitle{Graph contrastive learning with adaptive augmentation}. In \bibinfo{booktitle}{\emph{Proceedings of the Web Conference 2021}}. \bibinfo{pages}{2069--2080}.
\newblock


\bibitem[Zhu et~al\mbox{.}(2020a)]%
        {grace}
\bibfield{author}{\bibinfo{person}{Yanqiao Zhu}, \bibinfo{person}{Yichen Xu}, \bibinfo{person}{Feng Yu}, \bibinfo{person}{Qiang Liu}, \bibinfo{person}{Shu Wu}, {and} \bibinfo{person}{Liang Wang}.} \bibinfo{year}{2020}\natexlab{a}.
\newblock \showarticletitle{Deep graph contrastive representation learning}.
\newblock \bibinfo{journal}{\emph{arXiv preprint arXiv:2006.04131}} (\bibinfo{year}{2020}).
\newblock


\end{thebibliography}

\end{document}